\documentclass[10pt,twocolumn,letterpaper]{article}

\usepackage[pagenumbers]{cvpr} % To force page numbers, e.g. for an arXiv version

\definecolor{cvprblue}{rgb}{0.21,0.49,0.74}
\usepackage[pagebackref,breaklinks,colorlinks,allcolors=cvprblue]{hyperref}
\usepackage{multirow}
\newcommand{\zx}{\color{black}}

\def\paperID{} % *** Enter the Paper ID here
\def\confName{CVPR}
\def\confYear{2026}

\title{Heuristic-inspired Reasoning Priors Facilitate Data-Efficient Referring Object Detection}

\author{
{Xu Zhang\textsuperscript{1} \quad Zhe Chen\textsuperscript{2,\ensuremath{\dagger}} \quad Jing Zhang\textsuperscript{3} \quad Dacheng Tao\textsuperscript{4,\ensuremath{\dagger}}
}\\
\normalsize \textsuperscript{1}The University of Sydney, Australia \quad 
\normalsize \textsuperscript{2}La Trobe University, Australia \quad \\
\normalsize \textsuperscript{3}
Wuhan University, China \quad 
\normalsize \textsuperscript{4}
Nanyang Technological University, Singapore \\
\tt\footnotesize 
xzha0930@uni.sydney.edu.au \quad zhe.chen@latrobe.edu.au \quad \{jingzhang.cv, dacheng.tao\}@gmail.com}



\begin{document}
\maketitle
\footnotetext[2]{Corresponding authors.}
\begin{abstract}
Most referring object detection (ROD) models, especially the modern grounding detectors, are designed for data-rich conditions, yet many practical deployments, such as robotics, augmented reality, and other specialized domains, would face severe label scarcity. In such regimes, end-to-end grounding detectors need to learn spatial and semantic structure from scratch, wasting precious samples. We ask a simple question: \emph{Can explicit reasoning priors help models learn more efficiently when data is scarce?} To explore this, we first introduce a Data-efficient Referring Object Detection (De-ROD) task, which is a benchmark protocol for measuring ROD performance in low-data and few-shot settings. We then propose the HeROD (Heuristic-inspired ROD), a lightweight, model-agnostic framework that injects explicit, heuristic-inspired spatial and semantic reasoning priors, which are interpretable signals derived based on the referring phrase, into 3 stages of a modern DETR-style pipeline: proposal ranking, prediction fusion, and Hungarian matching. By biasing both training and inference toward plausible candidates, these priors promise to improve label efficiency and convergence performance. On RefCOCO, RefCOCO+, and RefCOCOg, HeROD consistently outperforms strong grounding baselines in scarce-label regimes. 
More broadly, our results suggest that integrating simple, interpretable reasoning priors provides a practical and extensible path toward better data-efficient vision–language understanding. Code at: https://github.com/xuzhang1199/HeROD. 
\end{abstract}    
\section{Introduction}
\label{sec:intro}

Referring Object Detection (ROD) \citep{kazemzadeh2014referitgame, yu2016modeling, mao2016generation} aims to localize the specific object in an image that matches a natural language description. While ROD and Referring Expression Comprehension (REC) are often used interchangeably, we use the term ROD in this paper.
Unlike generic object detection \citep{girshick2015fast, ren2015faster, redmon2016you, liu2016ssd, tian2019fcos, lin2017focal}, which identifies all instances from a fixed category set, ROD focuses on a single, linguistically specified target (e.g., “the person on the left wearing a red hat”). This requires not only object recognition but also fine-grained visual–semantic alignment and spatial reasoning. Recent advances in multimodal transformers \citep{vaswani2017attention} have significantly improved ROD performance. Models such as GLIP \citep{li2022grounded} and Grounding DINO \citep{liu2023grounding} leverage large-scale pre-training on image–text pairs to unify phrase grounding and object detection, achieving state-of-the-art results in data-rich scenarios.

\begin{figure}[t]
\centering
\centerline{\includegraphics[width=\columnwidth]{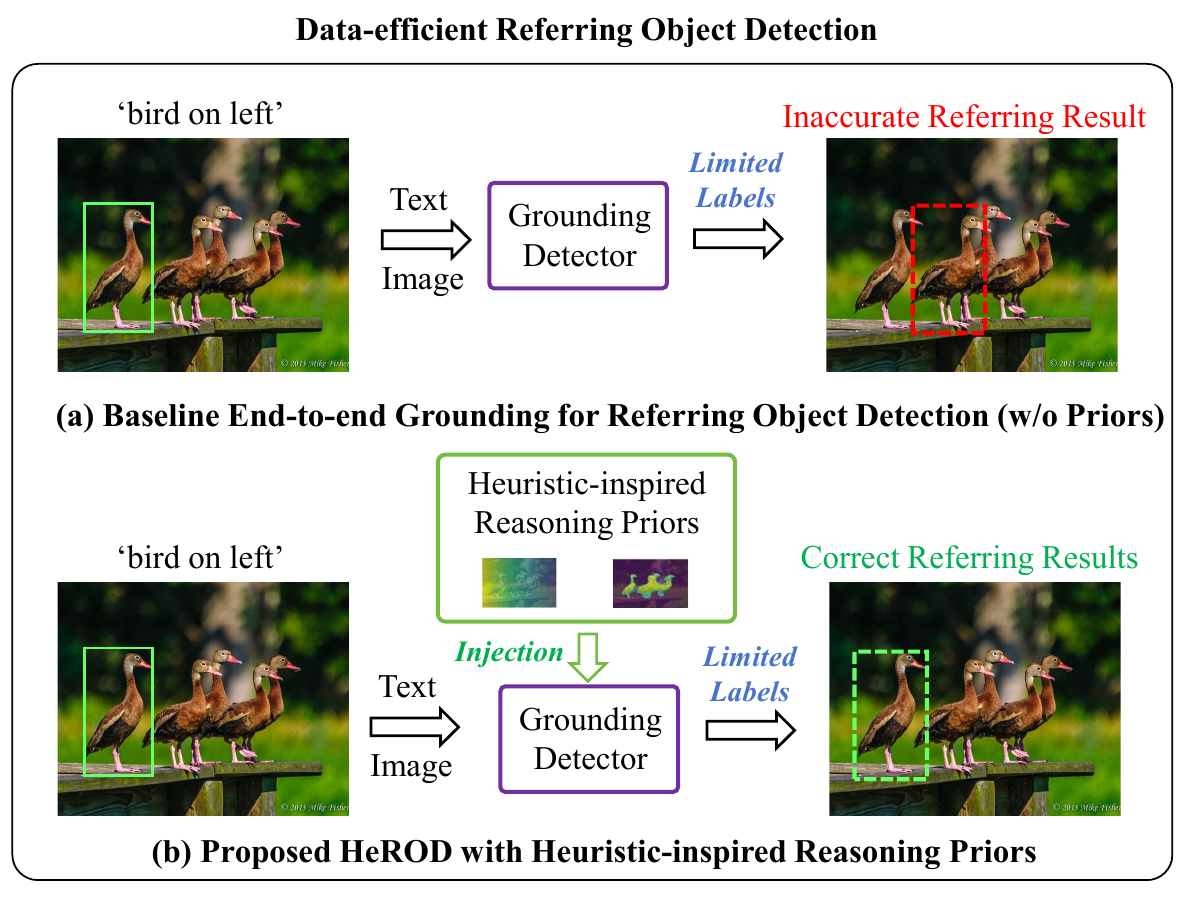}}
\caption{Conceptual illustration. (a) Without priors, grounding detectors trained with limited labels often ignore spatial/semantic cues (e.g., “on the left”). (b) HeROD injects heuristic-inspired reasoning priors (spatial + semantic), guiding the detector toward correct localization and improving data-efficient learning.
}
\label{title-fig}
% \end{center}
\end{figure}
Despite these advances, adapting ROD models to new domains or deployment scenarios remains challenging when annotated data are scarce. Many practical applications, such as robotics in unstructured environments or medical imaging with domain-specific terminology, cannot afford the scale of labeled data typically required for common pre-training or standard fine-tuning of related foundation models. In both low-data and full-data regimes, current grounding detectors primarily rely on end-to-end learning to implicitly acquire spatial relationships and visual–semantic associations relevant to the task. This would force the model to “rediscover” basic concepts (e.g., relative position, object attributes, inter-object relations) from task-specific examples. 
Under data-scarce conditions, this process can be highly sample-inefficient and prone to overfitting. While large-scale pre-training imparts broad vision–language alignment, fine-grained spatial cues, and complex attribute combinations, which are common in referring expressions, are often underrepresented \cite{chen2025revisiting}, leading to slower convergence and less precise localization when adapting with limited labels.

Recognizing these limitations, we first introduce Data-efficient Referring Object Detection (De-ROD), a tailored benchmarking protocol for systematically evaluating ROD models in low-data and few-shot settings. De-ROD emphasizes the need for models that generalize well with limited annotations, reflecting realistic deployment environments where large-scale, task-specific annotation is rarely feasible. We believe that this setting has been largely overlooked in existing ROD research and presents significant potential for real-world applications.
By addressing the De-ROD, our central insight is that the absence of explicit inductive biases leaves models to rely solely on implicit learning from scarce examples in De-ROD regimes. This motivates our research: to inject explicit reasoning signals (\emph{i.e.,} interpretable cues about spatial layout and semantic content) into the detection pipeline, so that models can focus on refining, rather than rediscovering, these fundamental relationships under limited supervision.

To this end, we propose \textbf{HeROD} (Heuristic-inspired Referring Object Detection), a lightweight, model-agnostic framework that reduces reliance on large training sets by injecting heuristic-inspired reasoning priors into the detection pipeline. In classic heuristic search methods such as A* \cite{hart1968formal}, search efficiency is greatly improved by heuristic costs that guide exploration toward promising candidates. Analogously, we define reasoning priors as explicit, interpretable spatial and semantic cues derived from the referring phrase and image, which bias both training and inference toward plausible regions. Prior work has shown that spatial cues are important for disambiguating referents \cite{yu2016modeling}, while semantic cues strongly influence grounding decisions \cite{qiao2020referring}. 
In HeROD, spatial reasoning priors are implemented as cardinal directions and simple composites of positional likelihood maps, and semantic reasoning priors are realized as text-conditioned visual scores from a pretrained vision–language model. These priors are injected into three phases of a modern DETR-style pipeline: proposal ranking, Hungarian matching, and final prediction fusion. By guiding the detector with reasoning priors, HeROD accelerates convergence, improves label efficiency, and lowers the risk of overfitting in scarce-label regimes. Empirically, HeROD achieves consistent gains in low-data and few-shot settings, while remaining competitive in full-data training. Figure~\ref{title-fig} conceptually illustrates how heuristic-inspired reasoning priors facilitate more data-efficient ROD.

\section{Related Work}
\noindent\textbf{Referring Object Detection and Grounding Models}
ROD extends object detection by requiring the localization of a target object described by a natural language expression. Unlike generic detectors \citep{girshick2015fast, ren2015faster, redmon2016you, liu2016ssd, tian2019fcos}, which predict all instances of predefined categories, ROD demands fine-grained visual–semantic alignment and spatial reasoning. Early REC/ROD approaches used modular attention mechanisms and multi-stage reasoning to process location, attribute, and relationship cues \citep{yu2016modeling, qiao2020referring}. The advent of transformers revolutionized detection: DETR \citep{carion2020end} framed detection as set prediction, eliminating anchors and NMS, with later variants improving convergence and representation quality through deformable attention \cite{zhu2020deformable}, conditional queries \cite{meng2021conditional}, dynamic anchor boxes \cite{liu2022dab}, denoising \cite{li2022dn}, and improved optimization \cite{chen2022recurrent}.
In multimodal detection, CLIP \citep{radford2021learning} demonstrated the power of large-scale vision–language pretraining for zero-shot classification, inspiring grounding-oriented detectors. MDETR \citep{kamath2021mdetr} and DQ-DETR \citep{liu2023dq} directly align text and image features in an end-to-end modulated detection framework, but still require extensive fine-tuning for ROD. GLIP \citep{li2022grounded} reformulates detection as phrase grounding, achieving strong zero-shot results but relying on the Dynamic Head \citep{dai2021dynamic}, which limits end-to-end fusion. UNINEXT \citep{yan2023universal} proposes a unified instance perception model, and Grounding DINO \citep{liu2023grounding} fuses text and image features at multiple transformer stages to achieve state-of-the-art zero-shot detection. However, as shown in \citep{liu2023grounding}, even these strong foundation detectors degrade sharply on ROD without large-scale task-specific fine-tuning, revealing their sensitivity to data availability.

\noindent\textbf{Data-Efficient and Few-Shot Object Detection}
Improving detection performance under limited annotation has been studied in the few-shot object detection (FSOD) literature. Meta-learning approaches such as FSRW \citep{kang2019few}, Meta-RCNN \citep{wu2020meta}, and FSDView \citep{xiao2022few} transfer category knowledge from base to novel classes. Later works improved adaptation through feature embedding and transfer learning, e.g., FSCE \citep{sun2021fsce}, UP-FSOD \citep{wu2021universal}, and T-GSEL \citep{zhang2025learning}. Another complementary direction is pretraining to improve generalization: DETReg \citep{bar2022detreg} uses unsupervised region priors with self-supervised feature encoders to reduce supervision needs. While these methods improve data efficiency in generic detection, ROD poses additional challenges due to the need for robust text comprehension and cross-modal alignment. To our knowledge, no prior work has defined a dedicated benchmark for evaluating ROD in low-data and few-shot regimes.

\noindent\textbf{Explicit Reasoning in Vision–Language Tasks}
Reasoning over spatial and semantic cues has been explored in REC/ROD through learned modules, such as MattNet’s location and relationship branches \citep{yu2018mattnet} or scene-graph grounding methods \citep{yang2019cross}. Some zero-shot or few-shot vision–language models incorporate external priors for improved grounding: TAS \citep{suo2023text}, for example, augments visual grounding with spatial heatmaps from text cues, while IterPrimE \citep{wang2025iterprime} iteratively refines attention maps using CLIP-based signals. These approaches, however, are typically designed for referring segmentation rather than detection, operate in purely zero-shot settings, or treat priors as post-processing rather than integrating them into training. In contrast, motivated by the importance of positional context \citep{yu2016modeling} and visual–semantic attributes \citep{qiao2020referring}, our method injects explicit spatial and semantic priors into detectors. 

%%%%%%%%%%%%%%%%%%%
\section{De-ROD Formulation}

To systematically evaluate a model's ability to adapt under data-scarce real-world applications, we introduce the \textbf{Data-efficient Referring Object Detection (De-ROD)} benchmarking protocol. De-ROD measures ROD performance in two realistic scenarios: \emph{Low-data ROD}, where training uses only a small percentage of the available dataset; and \emph{Few-shot ROD}, where the model must generalize to novel object categories with limited examples after training on abundant support categories. 
Formally, let $X_i$ and $y_i$ denote the $i$-th image–description pair and its corresponding annotation, respectively. The complete ROD dataset is 
$D = \{(X_i, y_i) \mid i = 1, \ldots, N\}$, where $N$ is the total number of annotated samples. We define subsets of $D$ to simulate different levels of annotation availability. Let $D_{De} = \{D_{fs}, D_{ld}\}$ denote the collection of all data-efficient subsets, where: \textbf{(1)} $D_{ld}$ (\emph{low-data subsets}) are formed by sampling a small percentage of the entire dataset without restricting to specific categories. \textbf{(2)} $D_{fs}$ (\emph{few-shot subsets}) contain a fixed number of novel-class examples and an abundant set of support-class examples.
We use $D^{(p)}_{sub}$ to denote a subset of $D$ containing $p$ samples.  
In the \emph{few-shot} setting, we follow common detection protocols \citep{kang2019few}, setting $p \in \{0.5\mathrm{k}, 1\mathrm{k}, 2\mathrm{k}\}$ for novel-class fine-tuning data, and $p \in \{1\mathrm{k}, 2\mathrm{k}\}$ for support-class training data.
This differs from typical few-shot learning for classification, which only uses a few examples per class. Since object detection involves accurate localization, it would require more samples for effective adaptation\citep{kang2019few}.
In the \emph{low-data} setting, following \citep{bar2022detreg}, $p$ represents a percentage of the full dataset. We consider $p \in \{0.1\%, 0.2\%, 0.5\%, 1\%, 2\%, 5\%\}$, covering a range from extremely scarce annotations to modestly scarce cases (5\%), where annotation is more feasible but still limited. These settings provide a controlled and comprehensive framework for evaluating label efficiency in ROD. Additional details are provided in \cref{Sec:dis}. It is worth mentioning that we also report the fully supervised results in the supplementary as a reference for overall robustness.
%%%%%%%%%%%%%%%%%%%

\begin{figure*}[t]
\centering
\centerline{\includegraphics[width=\linewidth]{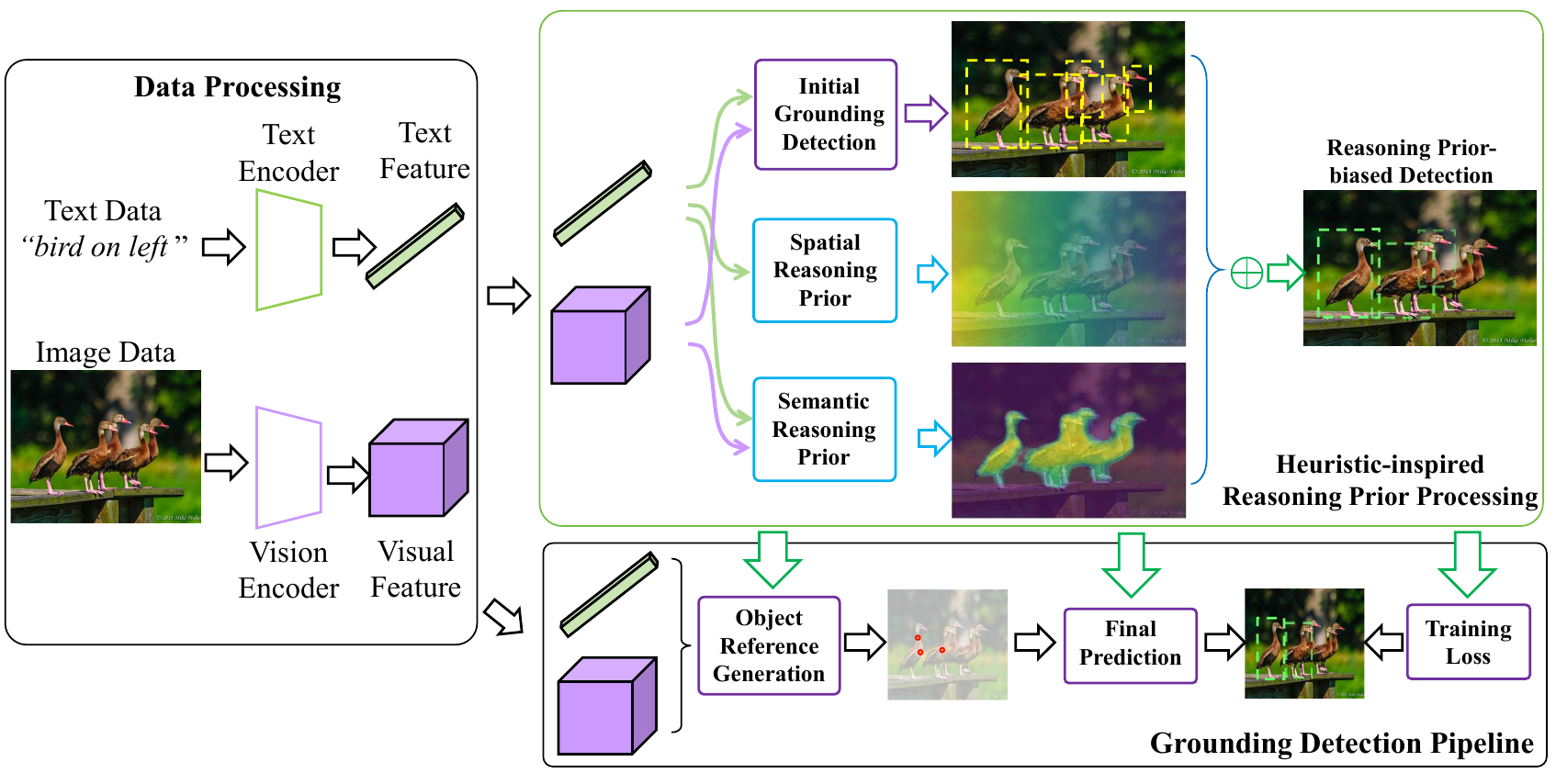}}
\caption{HeROD pipeline. Text and image features are encoded, and spatial/semantic reasoning priors are derived from the phrase and image. These priors are injected into reference generation, final prediction, and training loss, guiding the detector toward plausible regions for more data-efficient learning.}
\label{vis-method}
\end{figure*}

\section{Our Approach}

\textbf{HeROD Overview:}
To achieve effective performance under the De-ROD, we propose \textbf{HeROD} (\textit{Heuristic-inspired Referring Object Detection}), a learning framework that augments modern grounding detectors with explicit reasoning \emph{heuristics}. In our context, \textbf{heuristics} are interpretable signals that encode priors about where the referred object is likely to appear and how it should visually correspond to the description. These signals are derived directly from the referring phrase and the paired image, and they provide an inductive bias that complements the detector’s learned scores. The key idea is to bias both training and inference toward spatially and semantically plausible candidates, thereby improving label efficiency in data-scarce regimes.

Formally, given a data sample $X_i = (I_i, d_i)$ consisting of an image $I_i$ and a natural language description $d_i$, modern grounding detectors assign each object candidate $o_j \in O_i$ \footnote{$O_i$ is the full detected object set.} a learned matching probability $P(o_j|I_i, d_i)$. This score is computed entirely by the detector’s learned parameters, typically without explicit inductive bias toward spatial or semantic cues. Without rich supervision, the lack of such priors may cause the model to under-utilize important information in $d_i$ (e.g., “on the left,” “man in a blue shirt”) and converge slowly toward correct matches.
Inspired by heuristic search algorithms such as A* \citep{hart1968formal} that injects heuristic estimates to facilitate search, we introduce heuristic signals $H(o_j, I_i, d_i)$ to guide candidate selection. The refined selection process is:
\begin{align}
\overline{o}_i 
&= \arg\max_{o_j \in O_i} \; H(o_j, I_i, d_i) \oplus P(o_j|I_i, d_i),
\label{eq:a*}
\end{align}
where $H(\cdot)$ aggregates two complementary components: 
spatial heuristics $H_s(o_j, d_i)$ and 
visual heuristics $H_v(o_j, d_i, I_i)$:
\begin{equation}
H(o_j, I_i, d_i) = H_s(o_j, d_i) \oplus H_v(o_j, d_i, I_i).
\label{eq:heu}
\end{equation}
In the above 2 formulations, the operator $\oplus$ denotes the integration rule for heuristic priors and detector outputs. Its form varies across pipeline stages: in candidate generation, $\oplus$ reduces to an additive operation for efficiency; in final prediction, it is realized as a learnable weighting module; and in loss, it modifies the Hungarian matching cost to bias supervision.
Intuitively, spatial heuristics capture relative location cues (e.g., ``on the left,'' ``at the top'') that constrain where the object is likely to appear \citep{yu2016modeling}, while visual heuristics capture attribute- or relation-based cues (e.g., ``man with a hat,'' ``blue bag'') that strengthen text–image alignment \citep{qiao2020referring}.

An overview of our approach is illustrated in Fig.~\ref{vis-method}, and the following sections detail the construction of $H_s$ and $H_v$, the design of stage-specific integration rules $\oplus$, and their implementation within existing grounding detectors.

\textbf{Heuristic-inspired Spatial Prior}
\label{spatial-heu}
Spatial descriptions are among the most common and interpretable cues in referring expressions, yet they are usually underrepresented in current grounding detectors and difficult to learn from limited annotations. For example, phrases such as ``on the left'' provide strong positional constraints, but grounding detectors normally have to rediscover such concepts purely from rich labeled data. To address this inefficiency, we explicitly encode spatial priors as heuristic-inspired maps that can be directly injected into the detection process.
Formally, we define a vocabulary $\mathcal{T}$ of spatial descriptors (\emph{e.g.}, ``left,'' ``right,'' ``top,'' ``bottom,'' or composites like ``top left'' or ``bottom right''). Given a referring description $d_i$, we identify relevant spatial terms (denoted as $t_i$) by 
\[
t_i = Split(d_i) \cap \mathcal{T},
\]
where $Split$ is the operation that splits a string $d_i$. 
For instance, if $d_i =$ ``person on the left,'' then $t_i =$ ``left.'' 

After extraction, each $t_i$ will be linked to a pre-calculated score map $M_s(t_i)$ aligned to the image plane, where higher values indicate a greater likelihood of containing the referent. For example, the ``left'' map assigns higher scores to pixels closer to the left boundary, implemented via a linear or Gaussian decay across horizontal coordinates. Composite terms are handled by combining their base maps through averaging or weighted fusion.
Thus, given a candidate region $o_j$, we can compute its spatial prior by indexing the score map at the region center:
\begin{equation}
    H_s(o_j, d_i) = M_s(t_i)[loc(o_j)], 
    \label{eq:sp-sem}
\end{equation}
where $loc(o_j)$ is the center of $o_j$, and $M_s(t_i)[\cdot]$ returns the spatial relevance at a given image location. Intuitively, if $d_i$ mentions ``left'', objects nearer to the left edge of the image will receive higher prior values.

This heuristic-inspired prior provides two benefits. First, it introduces an explicit and interpretable inductive bias: the model is guided toward spatially plausible regions without requiring additional annotations. Second, it improves label efficiency in data-scarce regimes, since basic spatial reasoning (e.g., relative position) no longer needs to be re-learned from scratch. Visualizations of the constructed spatial prior maps are provided in \cref{vis-scoremap} of the supplementary materials.

%%%
\textbf{Heuristic-inspired Visual Prior}
\label{visual-heu}
Most referring expressions also specify visual attributes or object semantics (e.g., ``man with a hat,'', ``dog next to the bicycle''). Such cues are crucial for discriminating between multiple nearby candidate objects, yet grounding detectors typically learn them implicitly from data. Similar to learning spatial cues, under limited annotations, this implicit learning can be inefficient. To alleviate this, we introduce a heuristic-inspired \textbf{visual prior} $H_v$, which explicitly encodes the degree of visual–semantic alignment between each candidate object $o_j$ and the referring phrase $d_i$.
Formally, we compute:
\begin{equation}
    H_v(o_j, d_i, I_i) = {Align}(I_i, d_i; o_j),
    \label{eq:visual-sem}
\end{equation}
where ${Align}(\cdot)$ is a vision–language alignment function that measures how well the visual region corresponding to $o_j$ matches the description $d_i$ in the context of image $I_i$.

A straightforward way of implementation is to use CLIP \citep{radford2021learning} by cropping each region $o_j$ and computing its similarity with $d_i$, but this is inefficient in detection pipelines with many proposals, as it requires repeated forward passes and scales poorly with the number of candidates. Instead, we adopt CLIPSeg \citep{luddecke2022image}, which takes the entire image and text once and produces dense, text-conditioned relevance maps. We obtain $H_v(o_j,d_i,I_i)$ by averaging the CLIPSeg scores inside each candidate box. This can yield interpretable priors that guide evaluation and can scale efficiently across proposals, which are advantages over the typical CLIP-based models. 
When applying, since CLIPSeg is largely insensitive to spatial terms, we directly use raw input phrases without modification to prompt CLIPSeg. We would like to highlight that HeROD does more than add CLIPSeg as an ensemble. Raw CLIPSeg maps are coarse, and they alone cannot alter the learning dynamics of a grounding detector; thus, a naive fusion gives only marginal gains. Instead, our framework integrates $H_v$ into proposal ranking, matching, and prediction fusion, ensuring these signals shape both training and inference to improve data efficiency.

To sum up, the spatial prior $H_s$ and the visual prior $H_v$ supply complementary reasoning signals: $H_s$ restricts search space by position, while $H_v$ enforces semantic fidelity to the description. By explicitly injecting them into learned detection, HeROD achieves more efficient and robust adaptation in data-scarce regimes. Some extra justification about both priors are in \cref{Sec:dis} of the supplementary.

\subsection{HeROD in DETR Pipeline}
\label{HeROD-pipeline}

In a DETR-style pipeline, a Transformer translates visual and textual embeddings into detection outputs. HeROD is designed to be agnostic to different DETR variants and can be injected into three stages of this pipeline: \textbf{(1) Object reference generation stage}, which ranks candidate proposals to filter out irrelevant embeddings. This is foundational for efficient convergence \citep{zhu2020deformable, liu2023grounding}. Here, spatial and visual priors adjust candidate scores before the top-$N$ selection, steering early processing toward plausible regions. \textbf{(2) Final prediction stage}, which helps the detector assign probabilities $P(o_j|I_i, d_i)$ to each candidate. Here, we fuse these with $H(o_j, I_i, d_i)$ through a lightweight adaptive module, allowing the model to learn how to weight priors against network predictions for the final argmax selection. \textbf{(3) Training objectives} based on Hungarian matching. In loss functions, priors are incorporated into the cost function, biasing supervision toward spatially and semantically consistent matches. 
Each stage presents different design challenges. For instance, the reference stage relies on top-$N$ selection rather than strict argmax, so priors must guide without discarding potentially useful proposals. In contrast, the final prediction stage benefits from an adaptive $\oplus$ that balances heuristics and detector scores. Similarly, the matching loss requires a careful formulation to integrate priors without overwhelming learned signals. 

To address these differences, we employ stage-specific integration rules for $\oplus$ in Eq.~\eqref{eq:a*}, rather than a single static fusion. This adaptive design ensures that heuristic-inspired priors are injected effectively while preserving efficiency across the DETR pipeline.
  
\textbf{Prior-injected Object Reference Generation:} 
Rather than selecting the top-1 candidate, the reference generation stage samples the top-$N$ proposals ranked by confidence scores \citep{zhang2022dino}. This step is important for efficient convergence, since it determines which proposals are carried forward through the DETR pipeline. To guide this process, we inject our heuristic-inspired priors directly into the ranking. 
However, we found that it is non-trivial to design explicit supervision for this stage. As a result, we attempt to adopt a simple and efficient integration rule: $\oplus$ is directly instantiated as \emph{addition}. Thus, for each candidate, its spatial and visual prior scores are summed with the detector’s confidence, biasing the top-$N$ selection toward plausible regions without additional computational cost. Formally:
\begin{equation}
    \overline{O}_i' = TopN_{o_j \in O_i'} \big[\, P(o_j|I_i, d_i) + H_s(o_j, d_i) + H_v(o_j, d_i, I_i)\,\big],
    \label{eq:depr-ref}
\end{equation}
where $TopN$ returns the $N$ highest-scoring candidates from the initial set $O_i'$, and $\overline{O}_i'$ denotes the resulting reference set. This additive fusion maintains the efficiency of standard top-$N$ sampling while allowing priors to shape candidate generation from the earliest stage of the pipeline.

\textbf{Prior-injected Final Prediction:} 
While simple additive fusion like Eq. \ref{eq:depr-ref} could suffice for reference generation, the final prediction stage involves a single top-1 decision with direct supervision from ground-truth annotations. This makes it possible, and beneficial, to use a more expressive integration of priors and detector scores. 
To achieve this, we implement $\oplus$ as a lightweight learnable module. Specifically, we concatenate the spatial prior $H_s$, visual prior $H_v$, and detector confidence $P(o_j|I_i, d_i)$, and feed them into a small multi-layer perceptron (MLP). The MLP learns to adaptively balance these signals, allowing the model to adjust how strongly it relies on priors versus detector outputs. This design maximizes flexibility while keeping extra complexity minimal. Formally, we have the following:
\begin{equation}
z_j = MLP\!\left( Cat\!\big[ H_s(o_j, d_i),\, H_v(o_j, d_i, I_i),\, P(o_j|I_i, d_i) \big] \right),
\label{eq:mlp-score}
\end{equation}
where $Cat[\cdot]$ denotes concatenation and $MLP$ is a small feed-forward network. Then, the final prediction is:
\begin{equation}
\overline{o}_i = \arg\max_{o_j \in \overline{O}_i'} z_j.
\label{eq:mlp-select}
\end{equation}
By learning this fusion, the detector can adaptively calibrate its reliance on explicit priors, improving robustness in data-scarce settings without introducing heavy computation.

\begin{table*}[t]
  \centering
    \caption{Top-1 accuracy comparison on the RefCOCO/+/g datasets for the low-data ROD. $\Delta$ denotes the gain of HeROD over its baseline. } 
    \vskip -0.1in
  \label{lowdata-ref}
  \resizebox{\linewidth}{!}{
  \begin{tabular}{l|ccc|ccc|ccc|ccc|ccc|ccc}
    %RefCOCO
    \hline
    \multicolumn{19}{c}{\textbf{RefCOCO}} \\
    \hline
    \multicolumn{1}{c|}{} & \multicolumn{3}{c|}{0.1\% Data} & \multicolumn{3}{c|}{0.2\% Data} & \multicolumn{3}{c|}{0.5\% Data} & \multicolumn{3}{c|}{1\% Data} & \multicolumn{3}{c|}{2\% Data} & \multicolumn{3}{c}{5\% Data} \\
    \hline
    Method & val & testA & testB & val & testA & testB & val & testA & testB & val & testA & testB & val & testA & testB & val & testA & testB \\
    \hline
    UNINEXT & 17.75 & 22.03 & 14.27 & 21.53 & 23.23 & 16.80 & 24.05 & 27.24 & 18.61 & 31.04 & 34.35 & 27.77 & 46.24 & 49.67 & 39.73 & 54.30 & 57.24 & 51.40 \\
    HeROD-U (ours) & \textbf{25.60} & \textbf{33.20} & \textbf{18.47} & \textbf{31.59} & \textbf{36.45} & \textbf{25.26} & \textbf{35.75} & \textbf{41.24} & \textbf{32.11} & \textbf{47.89} & \textbf{53.51} & \textbf{40.12} & \textbf{59.45} & \textbf{65.21} & \textbf{52.99} & \textbf{65.33} & \textbf{70.53} & \textbf{58.74} \\
    $\Delta$&+7.85&+11.17&+4.20&+10.06&+13.22&+8.46&+11.70&+14.00&+13.5&+16.85&+19.16&+12.35&+13.21&+15.54&+13.26&+11.03&+13.29&+7.34 \\
    
    \hline
    Grounding DINO & 57.93 & 65.64 & 50.26 & 59.17 & 67.72 & 50.72 & 60.97 & 69.40 & 52.05 & 63.66 & 72.41 & 56.70 & 66.60 & 74.69 & 58.37 & 79.32 & 84.37 & 73.50 \\
    HeROD-G (ours) & \textbf{70.82} & \textbf{76.95} & \textbf{64.67} & \textbf{73.37} & \textbf{78.13} & \textbf{66.52} & \textbf{74.32} & \textbf{79.53} & \textbf{67.32} & \textbf{77.91} & \textbf{82.91} & \textbf{72.33} & \textbf{79.43} & \textbf{84.64} & \textbf{73.41} & \textbf{82.51} & \textbf{87.63} & \textbf{78.25} \\

    $\Delta$ & +12.89 & +11.31 & +14.41 & +14.20 & +10.41 & +15.80 & +13.35 & +10.13 & +15.27 & +14.25 & +10.50 & +15.63 & +12.83 & +9.95 & +15.04 & +3.19 & +3.26 & +4.75 \\
    
    \hline
    \hline
    \multicolumn{19}{c}{\textbf{RefCOCO+}} \\
    \hline
    \multicolumn{1}{c|}{} & \multicolumn{3}{c|}{0.1\% Data} & \multicolumn{3}{c|}{0.2\% Data} & \multicolumn{3}{c|}{0.5\% Data} & \multicolumn{3}{c|}{1\% Data} & \multicolumn{3}{c|}{2\% Data} & \multicolumn{3}{c}{5\% Data} \\
    \hline
    Method & val & testA & testB & val & testA & testB & val & testA & testB & val & testA & testB & val & testA & testB & val & testA & testB \\
    \hline
    UNINEXT & 16.60 & 20.36 & 12.38 & 19.68 & 23.72 & 14.60 & 20.59 & 24.50 & 15.44 & 26.10 & 28.50 & 22.05 & 26.85 & 30.42 & 23.83 & 34.51 & 37.25 & 29.13 \\
    HeROD-U (ours) & \textbf{27.03} & \textbf{35.10} & \textbf{15.50} & \textbf{30.27} & \textbf{39.36} & \textbf{19.92} & \textbf{34.17} & \textbf{46.61} & \textbf{22.85} & \textbf{38.68} & \textbf{48.74} & \textbf{27.55} & \textbf{42.92} & \textbf{53.60} & \textbf{30.21} & \textbf{51.27} & \textbf{60.25} & \textbf{39.78} \\
    $\Delta$ &+10.43&+14.74&+3.12&+10.59&+15.64 &+5.32&+13.58&+22.11&+7.41&+12.58&+20.24&+5.50&+16.07 &+23.18&+6.38&+16.76&+23.00&+10.65 \\
    
    \hline
    Grounding DINO & 61.68 & 69.56 & 51.05 & 63.30 & 72.35 & 52.16 & 66.32 & 74.55 & 54.12 & 68.24 & 76.93 & 56.33 & 69.95 & 78.89 & 58.87 & 71.90 & 80.86 & 62.00 \\
    
    HeROD-G (ours) & \textbf{62.52} & \textbf{70.17} & \textbf{51.52} & \textbf{65.27} & \textbf{74.07} & \textbf{53.04} & \textbf{66.98} & \textbf{75.13} & \textbf{55.14} & \textbf{69.30} & \textbf{77.80} & \textbf{57.07} & \textbf{70.98} & \textbf{79.53} & \textbf{59.24} & \textbf{72.60} & \textbf{81.07} & \textbf{63.08} \\

    $\Delta$ &+0.84 &+0.61 &+0.47 &+1.97 &+1.72 &+0.88 &+0.66 &+0.58 &+1.02 &+1.06 &+0.87 &+0.74 &+1.03 &+0.64 &+0.37 &+0.70 &+0.21 &+1.08 \\
    \hline
    \hline
    \multicolumn{19}{c}{\textbf{RefCOCOg}} \\
    \hline
    \multicolumn{1}{c|}{} & \multicolumn{3}{c|}{0.1\% Data} & \multicolumn{3}{c|}{0.2\% Data} & \multicolumn{3}{c|}{0.5\% Data} & \multicolumn{3}{c|}{1\% Data} & \multicolumn{3}{c|}{2\% Data} & \multicolumn{3}{c}{5\% Data} \\
    \hline
    Method & \multicolumn{3}{c|}{val \quad \quad \quad test} & \multicolumn{3}{c|}{val \quad \quad \quad test} & \multicolumn{3}{c|}{val \quad \quad \quad test} & \multicolumn{3}{c|}{val \quad \quad \quad test} & \multicolumn{3}{c|}{val \quad \quad \quad test} & \multicolumn{3}{c}{val \quad \quad \quad test} \\
    \hline
    UNINEXT & \multicolumn{3}{c|}{17.18 \quad \quad 16.69} & \multicolumn{3}{c|}{20.61 \quad \quad 21.60} & \multicolumn{3}{c|}{23.18 \quad \quad 23.38} & \multicolumn{3}{c|}{25.18 \quad \quad 24.92} & \multicolumn{3}{c|}{29.02 \quad \quad 28.67} & \multicolumn{3}{c}{37.77 \quad \quad 37.29} \\
    HeROD-U (ours) & \multicolumn{3}{c|}{\textbf{20.65} \quad \quad \textbf{20.53}} & \multicolumn{3}{c|}{\textbf{28.82} \quad \quad \textbf{28.41}} & \multicolumn{3}{c|}{\textbf{30.17} \quad \quad \textbf{31.73}} & \multicolumn{3}{c|}{\textbf{32.72} \quad \quad \textbf{33.24}} & \multicolumn{3}{c|}{\textbf{41.05} \quad \quad \textbf{41.06}} & \multicolumn{3}{c}{\textbf{51.00} \quad \quad \textbf{51.57}} \\

    $\Delta$ & \multicolumn{3}{c|}{+3.47 \quad \quad +3.84} & \multicolumn{3}{c|}{+8.21 \quad \quad+6.81}& \multicolumn{3}{c|}{+6.99 \quad \quad+8.35}& \multicolumn{3}{c|}{+7.54 \quad \quad+8.32}& \multicolumn{3}{c|}{+12.03 \quad \quad+12.39}& \multicolumn{3}{c}{+13.23 \quad \quad+14.28} \\			
    \hline
    Grounding DINO & \multicolumn{3}{c|}{70.59 \quad \quad 71.35} & \multicolumn{3}{c|}{72.24 \quad \quad 71.79} & \multicolumn{3}{c|}{72.55 \quad \quad 73.00} & \multicolumn{3}{c|}{73.92 \quad \quad 73.61} & \multicolumn{3}{c|}{74.47 \quad \quad 75.06} & \multicolumn{3}{c}{76.63 \quad \quad 77.24} \\
    HeROD-G (ours) & \multicolumn{3}{c|}{\textbf{72.20} \quad \quad \textbf{72.93}} & \multicolumn{3}{c|}{\textbf{73.96} \quad \quad \textbf{74.31}} & \multicolumn{3}{c|}{\textbf{75.22} \quad \quad \textbf{75.55}} & \multicolumn{3}{c|}{\textbf{76.04} \quad \quad \textbf{76.62}} & \multicolumn{3}{c|}{\textbf{77.21} \quad \quad \textbf{77.40}} & \multicolumn{3}{c}{\textbf{78.76} \quad \quad \textbf{79.07}} \\
    $\Delta$&  \multicolumn{3}{c|}{+1.61 \quad \quad +1.58} & \multicolumn{3}{c|}{+1.72 \quad \quad +2.52} & \multicolumn{3}{c|}{+2.67 \quad \quad +2.55} & \multicolumn{3}{c|}{+2.12 \quad \quad +3.01} & \multicolumn{3}{c|}{+2.74 \quad \quad +2.34} & \multicolumn{3}{c}{+2.13 \quad \quad +1.83} \\
    \hline
  \end{tabular}
  }

\end{table*}

\begin{table*}[t]

  \centering
    \caption{Top-1 accuracy comparison on the RefCOCO dataset for the few-shot ROD. }

  \label{fewshot-refcoco}
  \resizebox{\linewidth}{!}{
  \begin{tabular}{l|c|ccc||ccc|ccc|ccc}
    \hline
    \multirow{2}{*}{Method} & \multirow{2}{*}{Support Data} & \multicolumn{3}{c||}{Support Set Training} & \multicolumn{3}{c|}{2k Finetune} & \multicolumn{3}{c|}{1k Finetune} & \multicolumn{3}{c}{0.5k Finetune} \\
    \cline{3-14}
     &  & val & testA & testB & val & testA & testB & val & testA & testB & val & testA & testB \\
    \hline
    Grounding DINO & 2k & \multirow{2}{*}{61.18} & \multirow{2}{*}{70.66} & \multirow{2}{*}{52.23} & 67.23 & 72.30 & 63.85 & 62.98 & 69.56 & 58.43 & 62.08 & 68.84 & 56.84 \\
    HeROD-G (ours) & 2k &  &  &  & \textbf{78.17} \textsuperscript{+10.94} & \textbf{80.50} \textsuperscript{+8.20} & \textbf{75.62} \textsuperscript{+11.77} & \textbf{76.90} \textsuperscript{+13.92} & \textbf{80.15} \textsuperscript{+10.59} & \textbf{74.31} \textsuperscript{+15.88} & \textbf{75.08} \textsuperscript{+13.00}& \textbf{78.75} \textsuperscript{+9.91} & \textbf{71.03} \textsuperscript{+14.19} \\
    
    \hline
    Grounding DINO & 1k & \multirow{2}{*}{59.28} & \multirow{2}{*}{68.62} & \multirow{2}{*}{50.52} & 64.74 & 69.91 & 61.02 & 61.81 & 67.93 & 56.86 & 60.65 & 67.74 & 54.78 \\
    HeROD-G (ours) & 1k &  &  &  & \textbf{76.81}\textsuperscript{+12.07}  & \textbf{79.51} \textsuperscript{+9.60} & \textbf{75.39} \textsuperscript{+14.37}& \textbf{75.56}\textsuperscript{+13.75}& \textbf{78.27} \textsuperscript{+10.34}& \textbf{73.54} \textsuperscript{+16.68}& \textbf{72.85} \textsuperscript{+12.20}& \textbf{76.91} \textsuperscript{+9.17}& \textbf{70.30}\textsuperscript{+15.52} \\
    \hline
  \end{tabular}
  }
\end{table*}

\textbf{Prior-injected Training Objectives:} 
Beyond candidate generation and final prediction, we also inject priors into the learning objectives. Specifically, we modify the Hungarian matching process \citep{stewart2016end, carion2020end}, which assigns ground-truth objects to final predictions. In standard DETR, the matching cost combines classification, box regression, and GIoU terms, but they may be unreliable at the beginning of training, especially in data-scarce regimes where classification logits are noisy and box accuracies are weak. As a result, assignments may be unstable and hinder convergence.
To address this, we incorporate heuristic-inspired priors into the matching cost so that supervision is biased toward proposals consistent with spatial and semantic cues. The modified cost is:
\begin{equation}
     Cost_h = Cost_{cls} + Cost_{bbox} + Cost_{giou} - H(o_j, d_i, I_i),
    \label{eq:depr-cost-func}
\end{equation}
where $H(\cdot)$ is the unified spatial and visual priors as discussed before. The subtraction ensures that candidates better aligned with heuristic-inspired priors are preferred during matching, leading to more stable and semantically meaningful assignments early in training.
Once the assignment is established, we also supervise the detector with a loss that integrates heuristic-inspired confidence:
\begin{equation}
     L_h = L_{cls}(o_i, o^*_i) + L_{bbox}(o_i, o^*_i) + L_{conf}(o_i, o^*_i),
    \label{eq:depr-loss}
\end{equation}
where $o^*_i$ is the matched ground truth. Here, $L_{cls}$ and $L_{bbox}$ follow standard DETR practice, while $L_{conf}$ is a mean squared error loss between predicted confidence and the scores of heuristic-inspired priors, treated as soft labels. This encourages the model not only to predict the correct box but also to align its confidence with prior-informed plausibility. 

\textbf{Further Discussions:} {\zx Our novelty lies in (1) being the first to systematically define and study data-efficient ROD as a task, and (2) introducing a principled, interpretable framework for injecting explicit reasoning priors under severe label scarcity. The key insight is that making simple cues explicit and actionable can substantially improve learning efficiency, instead of relying on implicit end-to-end learning to rediscover them from limited data. 

}

\section{Experiments}
%%%%%%%%%%%%%%%%
\subsection{De-ROD Settings}
To rigorously evaluate how grounding detectors adapt to ROD under limited supervision, we introduce the \textbf{De-ROD} benchmark, which defines two primary data-scarce regimes: \emph{low-data ROD} and \emph{few-shot ROD}. These conditions, inspired by DETReg \citep{bar2022detreg} and FSRW \citep{kang2019few}, explicitly test learning efficiency when annotations are scarce. For completeness, we also report results in the fully supervised setting on the complete RefCOCO/+/g datasets \citep{yu2016modeling, mao2016generation}, which remains the standard evaluation protocol. 
This helps illustrate whether our method is not only effective in scarce-label regimes but also competitive when rich labels are available.

\textbf{Low-data ROD.}
We simulate extremely limited supervision by randomly sampling small percentages of the training data, specifically 0.1\%, 0.2\%, 0.5\%, 1\%, 2\%, and 5\%. Each larger subset is a superset of the smaller ones to ensure fairness. Models are evaluated on the standard validation and test sets: RefCOCO/+ (validation, testA, testB) and RefCOCOg (validation, test). This protocol reflects realistic deployment scenarios, ranging from extremely scarce annotations (0.1\%) to modestly available data (5\%).

\textbf{Few-shot ROD.}
We further consider generalization from abundant \emph{support} classes to scarce \emph{novel} classes. Motivated by the human-centric bias in RefCOCO (testA contains humans, testB contains other objects), we define humans as support classes ($\mathcal{C}_s$) and non-human categories as novel classes ($\mathcal{C}_n$). 
We construct support sets of 1k and 2k human pairs, and novel-class fine-tuning sets of 0.5k, 1k, and 2k non-human pairs, where larger sets subsume the smaller. Since novel classes have far fewer examples per category than humans, this qualifies as a few-shot regime. Evaluation follows RefCOCO’s official splits (testA for humans, testB for non-humans), with validation results also reported. This setup complements the low-data protocol by explicitly stressing cross-class generalization.
We select humans as the support class ($\mathcal{C}_s$) because RefCOCO contains abundant human annotations (testA), while non-human objects (testB) have fewer instances per class, making them a natural candidate for novel categories. This split reflects realistic deployment scenarios (e.g., robotics or AR), where systems often have rich supervision on frequently encountered classes (humans, hands) but must adapt with few samples to less common objects

\subsubsection{Benchmark Metrics and Baselines}
Following common practice in ROD, we report \textbf{top-1 accuracy}, which measures whether the highest-scoring prediction matches the referred object. 

For baselines, we adopt two strong and complementary detectors: 
(1) Grounding DINO \citep{liu2023grounding}, a widely recognized foundation detector with superior zero-shot detection ability, and 
(2) UNINEXT \citep{yan2023universal}, a recent universal instance perception model that achieves state-of-the-art ROD performance after fine-tuning on mixed datasets. 
We emphasize that to the best of our knowledge, no prior work has evaluated truly data-efficient ROD; thus, HeROD establishes the first benchmark in this regime. While newer detectors exist\cite{wu2024general}, they have not been developed and reported in low-data ROD. Meanwhile, cutting-edge grounding detectors are still mostly based on the architecture of Grounding DINO or UNINEXT, so we believe our comparisons with these two well-established foundations provide both fairness and reproducibility. 
For fairness, we instantiate HeROD on both baselines, denoted \textbf{HeROD-G} (with Grounding DINO) and \textbf{HeROD-U} (with UNINEXT). Grounding DINO and HeROD-G use Swin-T \citep{liu2021swin} as image encoder, BERT \citep{devlin2018bert} as text encoder, and DINO \citep{zhang2022dino} as detector. UNINEXT and HeROD-U use ResNet-50 \citep{he2016deep} and Deformable DETR \citep{zhu2020deformable}. All models are initialized with official pre-trained weights. 

Training and testing protocols are identical for baselines and HeROD variants. We train with AdamW \citep{loshchilov2017decoupled}, using learning rates of $1\mathrm{e}{-5}$ for the text encoder and $2\mathrm{e}{-5}$ for other modules, with a linear decay for the text encoder following MDETR \citep{kamath2021mdetr}. The image encoder is frozen to preserve its strong pretrained features. Low-data ROD training runs for 5 epochs (10 epochs for 0.1\%/0.2\% subsets) with a drop after the penultimate epoch; few-shot ROD involves 3 epochs on the support set followed by 3 epochs of novel-class fine-tuning. All experiments are conducted on the same hardware with a batch size of 8. 

\subsection{Detailed Results}
\subsubsection{Overall Results}
\textbf{Low-Data ROD.}  As shown in \cref{lowdata-ref}, HeROD consistently improves over the baseline detectors Grounding DINO and UNINEXT across all low-data regimes. Even at the extreme 0.1\% training split, HeROD delivers a gain of up to $\Delta=+12.9$ points on RefCOCO validation compared to Grounding DINO. The gains remain stable as the proportion of labeled data increases, confirming that the improvements are not isolated to the extreme regime but reflect a general advantage in label efficiency. These results indicate that our explicit priors bias the search space toward spatially and semantically plausible candidates, thereby accelerating convergence. The effect is analogous to heuristic search in A*, where guidance reduces exploration cost and leads to more efficient goal discovery. Importantly, the consistent $\Delta$ across RefCOCO/+/g demonstrates that HeROD provides a reliable mechanism for exploiting limited supervision.

{\zx 
On RefCOCO+, the relative improvements over Grounding DINO are less pronounced than those on RefCOCO/g. We attribute this to the design of RefCOCO+, which excludes absolute spatial cues from annotations \citep{yu2016modeling}. As a result, the spatial priors in HeROD have less opportunity to contribute, while the remaining visual descriptions are often generic and therefore more likely to be already captured by strongly pretrained detectors such as Grounding DINO, leaving limited room for further improvement. This also explains the different gain magnitudes across baselines: UNINEXT benefits more consistently, whereas strong visual-semantic alignment in Grounding DINO leads to smaller but still consistent gains. Nevertheless, HeROD still remains effective on RefCOCO+ and provides measurable value under extremely low-data settings where spatial guidance is less relevant and visual guidance is rough. We view this as motivation to extend HeROD with richer relational priors in future work.
}

\textbf{Few-Shot ROD.}  
Following the common protocol in \citep{kang2019few}, we first train all models on a support set (human category) before fine-tuning on limited samples from novel classes (non-human). Results are presented in \cref{fewshot-refcoco}. We primarily conduct evaluation using Grounding DINO for two reasons. First, Grounding DINO represents the most widely adopted foundation-level grounding detector. Second, while other detectors (e.g., UNINEXT) are competitive under full-data regimes, they are not explicitly optimized for few-shot transfer and lack standardized evaluation protocols for support vs. novel class splits. By anchoring the few-shot evaluation to Grounding DINO, we ensure both reproducibility and comparability.  Importantly, we emphasize that our contributions are \emph{orthogonal} to the choice of backbone. The HeROD framework is model-agnostic and could, in principle, be applied to other modern detectors or multimodal LLMs once consistent few-shot protocols are established. For now, Grounding DINO provides the most reliable testbed for analyzing few-shot behavior in ROD.

Looking at the results, since both baselines and HeROD share the same support pretraining, their performance is identical at that stage. After fine-tuning, the baselines achieve moderate gains on novel classes (testB) but suffer degradation on support classes (testA), especially with 0.5k or 1k novel samples. This reflects overfitting to scarce novel data, which disrupts previously learned support-class representations. In contrast, HeROD maintains balanced performance: it improves detection on novel classes while preserving accuracy on support classes. Across all settings, HeROD yields consistent $\Delta$ gains on both testA and testB, showing that our priors regularize adaptation by grounding the model in interpretable spatial and semantic signals. This allows the detector to adapt efficiently without catastrophic forgetting, a property that baseline detectors lack.

%%%%%%%%%%%%%

\begin{table}[t]
    \centering
    \small
    \caption{Ablation studies on the RefCOCO dataset for HeROD. $\mathcal{H}$ represents the heuristics. `Reference Gen' and `Learning Proc' represent Object Reference Generation and Learning Process.}
    \label{ablation-refcoco}
   \resizebox{\linewidth}{!}{
    \begin{tabular}{l|c|c|c}
    \hline
    Reference Gen & Final Prediction & Learning Proc & Performance \\
    \hline
    w/o $\mathcal{H}$ & Original          & w/o $\mathcal{H}$ & 63.66 \\
    \hline
    w/o $\mathcal{H}$ & Heuristic (static) & w/ $\mathcal{H}$  & 64.78\\
    \hline
    w/o $\mathcal{H}$ & Heuristic (adaptive) & w/ $\mathcal{H}$  & 71.33\\
    % \hline
    w/ $\mathcal{H}$  & Heuristic (adaptive) & w/ $\mathcal{H}$  & \textbf{77.91}\\
    \hline
    \end{tabular}%
    }
\end{table}

\begin{table}[t]

  \centering
  \caption{Ablation studies on the RefCOCO dataset for $H_s$, $H_v$ in different stages of HeROD.}
  \vskip -0.1in
  \label{ablation-Heu-refcoco}
  \resizebox{0.85\linewidth}{!}{
  \begin{tabular}{cc|cc|c}
    \hline
    \multicolumn{2}{c|}{Reference Generation} & \multicolumn{2}{c|}{Final Prediction} & \multirow{2}{*}{Performance} \\
    $H_s$ & $H_v$ & $H_s$ & $H_v$ & \\
    \hline
    \checkmark   & \checkmark  & \checkmark  &   & 76.36 \\
    % \hline
    \checkmark   & \checkmark  &  & \checkmark  & 74.93 \\
    \hline
    % w/o  & w/o & w/  & w/  & 71.33 \\
    \checkmark   &  & \checkmark  & \checkmark  & 76.02 \\
      & \checkmark  & \checkmark  & \checkmark  & 72.43 \\
    \hline
    \checkmark   & \checkmark  & \checkmark  & \checkmark  & \textbf{77.91} \\
    \hline
  \end{tabular}
  }

\end{table}

\subsubsection{Ablation Study}
\textbf{Pipeline Design.}  
To validate the effectiveness of each design choice in HeROD, we performed ablation studies on RefCOCO using only 1\% of the training data. Results are shown in \cref{ablation-refcoco}. We compare the baseline final prediction with two variants of our prior-injected final prediction: a \emph{static} variant that fuses heuristic-inspired priors by direct addition, and an \emph{adaptive} variant that employs a lightweight MLP to balance heuristics with detector scores under supervision. Both variants outperform the baseline, confirming that explicit priors improve data efficiency. Moreover, the adaptive variant achieves the strongest performance, highlighting the value of learnable fusion guided by our heuristic-augmented loss.

\textbf{Contribution of Each Stage.}  
We further isolate the impact of prior injection at different stages of the DETR pipeline. As shown in \cref{ablation-refcoco}, incorporating heuristics during reference generation yields clear gains by improving the quality of candidate proposals that enter later decoding layers. This effect demonstrates that early-stage biasing and late-stage adaptive balancing are complementary.

\textbf{Role of Spatial vs. Visual Priors.}  
Ablation studies on the spatial and visual priors are presented in \cref{ablation-Heu-refcoco}. Both components contribute positively: spatial priors help when positional cues are explicit in the phrase (e.g., “on the left”), while visual priors guide attribute-based matching (e.g., “man in a blue shirt”). Their combination provides the most consistent improvements, confirming our hypothesis that reasoning from both spatial layout and semantic attributes is necessary for robust data-efficient ROD.

\subsubsection{Qualitative and More Results.}  
We provide further ablations and visualizations in \cref{sec:further-abltation} and \cref{sec:visualization} in supplementary materials. 
These include (1) extended ablations on the impact of CLIPSeg integration and loss weighting, (2) evaluation under cleaned and corrected ROD datasets, and (3) qualitative examples showing how injected priors redirect model attention toward plausible regions, particularly in extremely low-data regimes. 

We also report results under full supervision in \cref{full-data-ref} and \cref{sec:data-rich} (supplementary). In this regime, HeROD remains competitive and consistently outperforms its baselines: on RefCOCO/+/g, HeROD-G improves Grounding DINO by $\Delta=+0.7$–$1.0$ points, and HeROD-U shows similar gains over UNINEXT. Compared with other ROD-specific detectors using backbones of similar complexity, HeROD achieves superior performance. Although the absolute improvements are smaller than in scarce-data regimes, these results confirm that prior injection does benefit learning when abundant annotations are available. Further discussions (e.g., extensibility, limitations) are in \cref{Sec:dis} (supp).

 \section{Conclusion}
We introduced \textbf{De-ROD}, the new benchmarking protocol for evaluating ROD under data scarcity, and proposed \textbf{HeROD}, a heuristic-inspired framework that injects explicit spatial and semantic priors into DETR-style detectors. By guiding both training and inference with interpretable priors, HeROD improves label efficiency in low-data regimes. Our results demonstrate that reasoning before learning offers a practical path toward robust, data-efficient ROD.

\textbf{Acknowledgement} Mr Xu Zhang was supported by Australian Research Council Projects in part by FL170100117. Dr Dacheng Tao’s research is partially supported by NTU RSR and Start Up Grants. Dr Jing Zhang was supported by the New Generation Artificial Intelligence-National Science and Technology Major Project (2025ZD0123602).
{
    \small
    \bibliographystyle{ieeenat_fullname}
    \bibliography{main}

\begin{thebibliography}{54}
\providecommand{\natexlab}[1]{#1}
\providecommand{\url}[1]{\texttt{#1}}
\expandafter\ifx\csname urlstyle\endcsname\relax
  \providecommand{\doi}[1]{doi: #1}\else
  \providecommand{\doi}{doi: \begingroup \urlstyle{rm}\Url}\fi

\bibitem[Bar et~al.(2022)Bar, Wang, Kantorov, Reed, Herzig, Chechik, Rohrbach, Darrell, and Globerson]{bar2022detreg}
Amir Bar, Xin Wang, Vadim Kantorov, Colorado~J Reed, Roei Herzig, Gal Chechik, Anna Rohrbach, Trevor Darrell, and Amir Globerson.
\newblock Detreg: Unsupervised pretraining with region priors for object detection.
\newblock In \emph{Proceedings of the IEEE/CVF conference on computer vision and pattern recognition}, pages 14605--14615, 2022.

\bibitem[Carion et~al.(2020)Carion, Massa, Synnaeve, Usunier, Kirillov, and Zagoruyko]{carion2020end}
Nicolas Carion, Francisco Massa, Gabriel Synnaeve, Nicolas Usunier, Alexander Kirillov, and Sergey Zagoruyko.
\newblock End-to-end object detection with transformers.
\newblock In \emph{European conference on computer vision}, pages 213--229. Springer, 2020.

\bibitem[Chen et~al.(2025{\natexlab{a}})Chen, Wei, Zhao, Song, Wu, Peng, Chan, and Zhang]{chen2024revisiting}
Jierun Chen, Fangyun Wei, Jinjing Zhao, Sizhe Song, Bohuai Wu, Zhuoxuan Peng, S-H~Gary Chan, and Hongyang Zhang.
\newblock Revisiting referring expression comprehension evaluation in the era of large multimodal models.
\newblock In \emph{Proceedings of the IEEE/CVF Conference on Computer Vision and Pattern Recognition}, pages 513--524, 2025{\natexlab{a}}.

\bibitem[Chen et~al.(2025{\natexlab{b}})Chen, Wei, Zhao, Song, Wu, Peng, Chan, and Zhang]{chen2025revisiting}
Jierun Chen, Fangyun Wei, Jinjing Zhao, Sizhe Song, Bohuai Wu, Zhuoxuan Peng, S-H~Gary Chan, and Hongyang Zhang.
\newblock Revisiting referring expression comprehension evaluation in the era of large multimodal models.
\newblock In \emph{Proceedings of the IEEE/CVF Conference on Computer Vision and Pattern Recognition}, pages 513--524, 2025{\natexlab{b}}.

\bibitem[Chen et~al.(2022)Chen, Zhang, and Tao]{chen2022recurrent}
Zhe Chen, Jing Zhang, and Dacheng Tao.
\newblock Recurrent glimpse-based decoder for detection with transformer.
\newblock In \emph{Proceedings of the IEEE/CVF conference on computer vision and pattern recognition}, pages 5260--5269, 2022.

\bibitem[Dai et~al.(2021)Dai, Chen, Xiao, Chen, Liu, Yuan, and Zhang]{dai2021dynamic}
Xiyang Dai, Yinpeng Chen, Bin Xiao, Dongdong Chen, Mengchen Liu, Lu Yuan, and Lei Zhang.
\newblock Dynamic head: Unifying object detection heads with attentions.
\newblock In \emph{Proceedings of the IEEE/CVF conference on computer vision and pattern recognition}, pages 7373--7382, 2021.

\bibitem[Deng et~al.(2021)Deng, Yang, Chen, Zhou, and Li]{deng2021transvg}
Jiajun Deng, Zhengyuan Yang, Tianlang Chen, Wengang Zhou, and Houqiang Li.
\newblock Transvg: End-to-end visual grounding with transformers.
\newblock In \emph{Proceedings of the IEEE/CVF international conference on computer vision}, pages 1769--1779, 2021.

\bibitem[Devlin et~al.(2019)Devlin, Chang, Lee, and Toutanova]{devlin2018bert}
Jacob Devlin, Ming-Wei Chang, Kenton Lee, and Kristina Toutanova.
\newblock Bert: Pre-training of deep bidirectional transformers for language understanding.
\newblock In \emph{Proceedings of the 2019 conference of the North American chapter of the association for computational linguistics: human language technologies, volume 1 (long and short papers)}, pages 4171--4186, 2019.

\bibitem[Du et~al.(2022)Du, Fu, Liu, and Wang]{du2022visualgroundingtransformers}
Ye Du, Zehua Fu, Qingjie Liu, and Yunhong Wang.
\newblock Visual grounding with transformers, 2022.

\bibitem[Gan et~al.(2020)Gan, Chen, Li, Zhu, Cheng, and Liu]{gan2020large}
Zhe Gan, Yen-Chun Chen, Linjie Li, Chen Zhu, Yu Cheng, and Jingjing Liu.
\newblock Large-scale adversarial training for vision-and-language representation learning.
\newblock \emph{Advances in Neural Information Processing Systems}, 33:\penalty0 6616--6628, 2020.

\bibitem[Girshick(2015)]{girshick2015fast}
Ross Girshick.
\newblock Fast r-cnn.
\newblock In \emph{Proceedings of the IEEE international conference on computer vision}, pages 1440--1448, 2015.

\bibitem[Hart et~al.(1968)Hart, Nilsson, and Raphael]{hart1968formal}
Peter~E Hart, Nils~J Nilsson, and Bertram Raphael.
\newblock A formal basis for the heuristic determination of minimum cost paths.
\newblock \emph{IEEE transactions on Systems Science and Cybernetics}, 4\penalty0 (2):\penalty0 100--107, 1968.

\bibitem[He et~al.(2016)He, Zhang, Ren, and Sun]{he2016deep}
Kaiming He, Xiangyu Zhang, Shaoqing Ren, and Jian Sun.
\newblock Deep residual learning for image recognition.
\newblock In \emph{Proceedings of the IEEE conference on computer vision and pattern recognition}, pages 770--778, 2016.

\bibitem[Kamath et~al.(2021)Kamath, Singh, LeCun, Synnaeve, Misra, and Carion]{kamath2021mdetr}
Aishwarya Kamath, Mannat Singh, Yann LeCun, Gabriel Synnaeve, Ishan Misra, and Nicolas Carion.
\newblock Mdetr-modulated detection for end-to-end multi-modal understanding.
\newblock In \emph{Proceedings of the IEEE/CVF international conference on computer vision}, pages 1780--1790, 2021.

\bibitem[Kang et~al.(2019)Kang, Liu, Wang, Yu, Feng, and Darrell]{kang2019few}
Bingyi Kang, Zhuang Liu, Xin Wang, Fisher Yu, Jiashi Feng, and Trevor Darrell.
\newblock Few-shot object detection via feature reweighting.
\newblock In \emph{Proceedings of the IEEE/CVF international conference on computer vision}, pages 8420--8429, 2019.

\bibitem[Kazemzadeh et~al.(2014)Kazemzadeh, Ordonez, Matten, and Berg]{kazemzadeh2014referitgame}
Sahar Kazemzadeh, Vicente Ordonez, Mark Matten, and Tamara Berg.
\newblock Referitgame: Referring to objects in photographs of natural scenes.
\newblock In \emph{Proceedings of the 2014 conference on empirical methods in natural language processing (EMNLP)}, pages 787--798, 2014.

\bibitem[Ke et~al.(2023)Ke, Ye, Danelljan, Tai, Tang, Yu, et~al.]{ke2023segment}
Lei Ke, Mingqiao Ye, Martin Danelljan, Yu-Wing Tai, Chi-Keung Tang, Fisher Yu, et~al.
\newblock Segment anything in high quality.
\newblock \emph{Advances in Neural Information Processing Systems}, 36:\penalty0 29914--29934, 2023.

\bibitem[Kingma and Ba(2014)]{kingma2014adam}
Diederik~P Kingma and Jimmy Ba.
\newblock Adam: A method for stochastic optimization.
\newblock \emph{arXiv preprint arXiv:1412.6980}, 2014.

\bibitem[Kirillov et~al.(2023)Kirillov, Mintun, Ravi, Mao, Rolland, Gustafson, Xiao, Whitehead, Berg, Lo, et~al.]{kirillov2023segment}
Alexander Kirillov, Eric Mintun, Nikhila Ravi, Hanzi Mao, Chloe Rolland, Laura Gustafson, Tete Xiao, Spencer Whitehead, Alexander~C Berg, Wan-Yen Lo, et~al.
\newblock Segment anything.
\newblock In \emph{Proceedings of the IEEE/CVF international conference on computer vision}, pages 4015--4026, 2023.

\bibitem[Li et~al.(2022{\natexlab{a}})Li, Zhang, Liu, Guo, Ni, and Zhang]{li2022dn}
Feng Li, Hao Zhang, Shilong Liu, Jian Guo, Lionel~M Ni, and Lei Zhang.
\newblock Dn-detr: Accelerate detr training by introducing query denoising.
\newblock In \emph{Proceedings of the IEEE/CVF conference on computer vision and pattern recognition}, pages 13619--13627, 2022{\natexlab{a}}.

\bibitem[Li et~al.(2022{\natexlab{b}})Li, Zhang, Zhang, Yang, Li, Zhong, Wang, Yuan, Zhang, Hwang, et~al.]{li2022grounded}
Liunian~Harold Li, Pengchuan Zhang, Haotian Zhang, Jianwei Yang, Chunyuan Li, Yiwu Zhong, Lijuan Wang, Lu Yuan, Lei Zhang, Jenq-Neng Hwang, et~al.
\newblock Grounded language-image pre-training.
\newblock In \emph{Proceedings of the IEEE/CVF conference on computer vision and pattern recognition}, pages 10965--10975, 2022{\natexlab{b}}.

\bibitem[Li and Sigal(2021)]{li2021referring}
Muchen Li and Leonid Sigal.
\newblock Referring transformer: A one-step approach to multi-task visual grounding.
\newblock \emph{Advances in neural information processing systems}, 34:\penalty0 19652--19664, 2021.

\bibitem[Lin et~al.(2017)Lin, Goyal, Girshick, He, and Doll{\'a}r]{lin2017focal}
Tsung-Yi Lin, Priya Goyal, Ross Girshick, Kaiming He, and Piotr Doll{\'a}r.
\newblock Focal loss for dense object detection.
\newblock In \emph{Proceedings of the IEEE international conference on computer vision}, pages 2980--2988, 2017.

\bibitem[Liu et~al.(2022)Liu, Li, Zhang, Yang, Qi, Su, Zhu, and Zhang]{liu2022dab}
Shilong Liu, Feng Li, Hao Zhang, Xiao Yang, Xianbiao Qi, Hang Su, Jun Zhu, and Lei Zhang.
\newblock Dab-detr: Dynamic anchor boxes are better queries for detr.
\newblock \emph{arXiv preprint arXiv:2201.12329}, 2022.

\bibitem[Liu et~al.(2023)Liu, Huang, Li, Zhang, Liang, Su, Zhu, and Zhang]{liu2023dq}
Shilong Liu, Shijia Huang, Feng Li, Hao Zhang, Yaoyuan Liang, Hang Su, Jun Zhu, and Lei Zhang.
\newblock Dq-detr: Dual query detection transformer for phrase extraction and grounding.
\newblock In \emph{Proceedings of the AAAI Conference on Artificial Intelligence}, pages 1728--1736, 2023.

\bibitem[Liu et~al.(2024)Liu, Zeng, Ren, Li, Zhang, Yang, Jiang, Li, Yang, Su, et~al.]{liu2023grounding}
Shilong Liu, Zhaoyang Zeng, Tianhe Ren, Feng Li, Hao Zhang, Jie Yang, Qing Jiang, Chunyuan Li, Jianwei Yang, Hang Su, et~al.
\newblock Grounding dino: Marrying dino with grounded pre-training for open-set object detection.
\newblock In \emph{European conference on computer vision}, pages 38--55. Springer, 2024.

\bibitem[Liu et~al.(2016)Liu, Anguelov, Erhan, Szegedy, Reed, Fu, and Berg]{liu2016ssd}
Wei Liu, Dragomir Anguelov, Dumitru Erhan, Christian Szegedy, Scott Reed, Cheng-Yang Fu, and Alexander~C Berg.
\newblock Ssd: Single shot multibox detector.
\newblock In \emph{European conference on computer vision}, pages 21--37. Springer, 2016.

\bibitem[Liu et~al.(2021)Liu, Lin, Cao, Hu, Wei, Zhang, Lin, and Guo]{liu2021swin}
Ze Liu, Yutong Lin, Yue Cao, Han Hu, Yixuan Wei, Zheng Zhang, Stephen Lin, and Baining Guo.
\newblock Swin transformer: Hierarchical vision transformer using shifted windows.
\newblock In \emph{Proceedings of the IEEE/CVF international conference on computer vision}, pages 10012--10022, 2021.

\bibitem[Loshchilov and Hutter(2017)]{loshchilov2017decoupled}
Ilya Loshchilov and Frank Hutter.
\newblock Decoupled weight decay regularization.
\newblock \emph{arXiv preprint arXiv:1711.05101}, 2017.

\bibitem[L{\"u}ddecke and Ecker(2022)]{luddecke2022image}
Timo L{\"u}ddecke and Alexander Ecker.
\newblock Image segmentation using text and image prompts.
\newblock In \emph{Proceedings of the IEEE/CVF conference on computer vision and pattern recognition}, pages 7086--7096, 2022.

\bibitem[Mao et~al.(2016)Mao, Huang, Toshev, Camburu, Yuille, and Murphy]{mao2016generation}
Junhua Mao, Jonathan Huang, Alexander Toshev, Oana Camburu, Alan~L Yuille, and Kevin Murphy.
\newblock Generation and comprehension of unambiguous object descriptions.
\newblock In \emph{Proceedings of the IEEE conference on computer vision and pattern recognition}, pages 11--20, 2016.

\bibitem[Meng et~al.(2021)Meng, Chen, Fan, Zeng, Li, Yuan, Sun, and Wang]{meng2021conditional}
Depu Meng, Xiaokang Chen, Zejia Fan, Gang Zeng, Houqiang Li, Yuhui Yuan, Lei Sun, and Jingdong Wang.
\newblock Conditional detr for fast training convergence.
\newblock In \emph{Proceedings of the IEEE/CVF international conference on computer vision}, pages 3651--3660, 2021.

\bibitem[Qiao et~al.(2020)Qiao, Deng, and Wu]{qiao2020referring}
Yanyuan Qiao, Chaorui Deng, and Qi Wu.
\newblock Referring expression comprehension: A survey of methods and datasets.
\newblock \emph{IEEE Transactions on Multimedia}, 23:\penalty0 4426--4440, 2020.

\bibitem[Radford et~al.(2021)Radford, Kim, Hallacy, Ramesh, Goh, Agarwal, Sastry, Askell, Mishkin, Clark, et~al.]{radford2021learning}
Alec Radford, Jong~Wook Kim, Chris Hallacy, Aditya Ramesh, Gabriel Goh, Sandhini Agarwal, Girish Sastry, Amanda Askell, Pamela Mishkin, Jack Clark, et~al.
\newblock Learning transferable visual models from natural language supervision.
\newblock In \emph{International conference on machine learning}, pages 8748--8763. PmLR, 2021.

\bibitem[Redmon et~al.(2016)Redmon, Divvala, Girshick, and Farhadi]{redmon2016you}
Joseph Redmon, Santosh Divvala, Ross Girshick, and Ali Farhadi.
\newblock You only look once: Unified, real-time object detection.
\newblock In \emph{Proceedings of the IEEE conference on computer vision and pattern recognition}, pages 779--788, 2016.

\bibitem[Ren et~al.(2015)Ren, He, Girshick, and Sun]{ren2015faster}
Shaoqing Ren, Kaiming He, Ross Girshick, and Jian Sun.
\newblock Faster r-cnn: Towards real-time object detection with region proposal networks.
\newblock \emph{Advances in neural information processing systems}, 28, 2015.

\bibitem[Stewart et~al.(2016)Stewart, Andriluka, and Ng]{stewart2016end}
Russell Stewart, Mykhaylo Andriluka, and Andrew~Y Ng.
\newblock End-to-end people detection in crowded scenes.
\newblock In \emph{Proceedings of the IEEE conference on computer vision and pattern recognition}, pages 2325--2333, 2016.

\bibitem[Sun et~al.(2021)Sun, Li, Cai, Yuan, and Zhang]{sun2021fsce}
Bo Sun, Banghuai Li, Shengcai Cai, Ye Yuan, and Chi Zhang.
\newblock Fsce: Few-shot object detection via contrastive proposal encoding.
\newblock In \emph{Proceedings of the IEEE/CVF conference on computer vision and pattern recognition}, pages 7352--7362, 2021.

\bibitem[Suo et~al.(2023)Suo, Zhu, and Yang]{suo2023text}
Yucheng Suo, Linchao Zhu, and Yi Yang.
\newblock Text augmented spatial aware zero-shot referring image segmentation.
\newblock In \emph{Findings of the Association for Computational Linguistics: EMNLP 2023}, pages 1032--1043, 2023.

\bibitem[Tian et~al.(2019)Tian, Shen, Chen, and He]{tian2019fcos}
Zhi Tian, Chunhua Shen, Hao Chen, and Tong He.
\newblock Fcos: Fully convolutional one-stage object detection.
\newblock In \emph{Proceedings of the IEEE/CVF international conference on computer vision}, pages 9627--9636, 2019.

\bibitem[Vaswani et~al.(2017)Vaswani, Shazeer, Parmar, Uszkoreit, Jones, Gomez, Kaiser, and Polosukhin]{vaswani2017attention}
Ashish Vaswani, Noam Shazeer, Niki Parmar, Jakob Uszkoreit, Llion Jones, Aidan~N Gomez, {\L}ukasz Kaiser, and Illia Polosukhin.
\newblock Attention is all you need.
\newblock \emph{Advances in neural information processing systems}, 30, 2017.

\bibitem[Wang et~al.(2025)Wang, Ni, Liu, Yuan, and Tang]{wang2025iterprime}
Yuji Wang, Jingchen Ni, Yong Liu, Chun Yuan, and Yansong Tang.
\newblock Iterprime: Zero-shot referring image segmentation with iterative grad-cam refinement and primary word emphasis.
\newblock In \emph{Proceedings of the AAAI Conference on Artificial Intelligence}, pages 8159--8168, 2025.

\bibitem[Wu et~al.(2021)Wu, Han, Zhu, and Yang]{wu2021universal}
Aming Wu, Yahong Han, Linchao Zhu, and Yi Yang.
\newblock Universal-prototype enhancing for few-shot object detection.
\newblock In \emph{Proceedings of the IEEE/CVF international conference on computer vision}, pages 9567--9576, 2021.

\bibitem[Wu et~al.(2024)Wu, Jiang, Liu, Yuan, Bai, and Bai]{wu2024general}
Junfeng Wu, Yi Jiang, Qihao Liu, Zehuan Yuan, Xiang Bai, and Song Bai.
\newblock General object foundation model for images and videos at scale.
\newblock In \emph{Proceedings of the IEEE/CVF Conference on Computer Vision and Pattern Recognition}, pages 3783--3795, 2024.

\bibitem[Wu et~al.(2020)Wu, Sahoo, and Hoi]{wu2020meta}
Xiongwei Wu, Doyen Sahoo, and Steven Hoi.
\newblock Meta-rcnn: Meta learning for few-shot object detection.
\newblock In \emph{Proceedings of the 28th ACM international conference on multimedia}, pages 1679--1687, 2020.

\bibitem[Xiao et~al.(2022)Xiao, Lepetit, and Marlet]{xiao2022few}
Yang Xiao, Vincent Lepetit, and Renaud Marlet.
\newblock Few-shot object detection and viewpoint estimation for objects in the wild.
\newblock \emph{IEEE transactions on pattern analysis and machine intelligence}, 45\penalty0 (3):\penalty0 3090--3106, 2022.

\bibitem[Yan et~al.(2023)Yan, Jiang, Wu, Wang, Luo, Yuan, and Lu]{yan2023universal}
Bin Yan, Yi Jiang, Jiannan Wu, Dong Wang, Ping Luo, Zehuan Yuan, and Huchuan Lu.
\newblock Universal instance perception as object discovery and retrieval.
\newblock In \emph{Proceedings of the IEEE/CVF Conference on Computer Vision and Pattern Recognition}, pages 15325--15336, 2023.

\bibitem[Yang et~al.(2019)Yang, Li, and Yu]{yang2019cross}
Sibei Yang, Guanbin Li, and Yizhou Yu.
\newblock Cross-modal relationship inference for grounding referring expressions.
\newblock In \emph{Proceedings of the IEEE/CVF conference on computer vision and pattern recognition}, pages 4145--4154, 2019.

\bibitem[Yu et~al.(2016)Yu, Poirson, Yang, Berg, and Berg]{yu2016modeling}
Licheng Yu, Patrick Poirson, Shan Yang, Alexander~C Berg, and Tamara~L Berg.
\newblock Modeling context in referring expressions.
\newblock In \emph{European conference on computer vision}, pages 69--85. Springer, 2016.

\bibitem[Yu et~al.(2018)Yu, Lin, Shen, Yang, Lu, Bansal, and Berg]{yu2018mattnet}
Licheng Yu, Zhe Lin, Xiaohui Shen, Jimei Yang, Xin Lu, Mohit Bansal, and Tamara~L Berg.
\newblock Mattnet: Modular attention network for referring expression comprehension.
\newblock In \emph{Proceedings of the IEEE conference on computer vision and pattern recognition}, pages 1307--1315, 2018.

\bibitem[Yu et~al.(2023)Yu, Seo, and Son]{yu2023zero}
Seonghoon Yu, Paul~Hongsuck Seo, and Jeany Son.
\newblock Zero-shot referring image segmentation with global-local context features.
\newblock In \emph{Proceedings of the IEEE/CVF conference on computer vision and pattern recognition}, pages 19456--19465, 2023.

\bibitem[Zhang et~al.(2022)Zhang, Li, Liu, Zhang, Su, Zhu, Ni, and Shum]{zhang2022dino}
Hao Zhang, Feng Li, Shilong Liu, Lei Zhang, Hang Su, Jun Zhu, Lionel~M Ni, and Heung-Yeung Shum.
\newblock Dino: Detr with improved denoising anchor boxes for end-to-end object detection.
\newblock \emph{arXiv preprint arXiv:2203.03605}, 2022.

\bibitem[Zhang et~al.(2025)Zhang, Chen, Zhang, Liu, and Tao]{zhang2025learning}
Xu Zhang, Zhe Chen, Jing Zhang, Tongliang Liu, and Dacheng Tao.
\newblock Learning general and specific embedding with transformer for few-shot object detection.
\newblock \emph{International Journal of Computer Vision}, 133\penalty0 (2):\penalty0 968--984, 2025.

\bibitem[Zhu et~al.(2020)Zhu, Su, Lu, Li, Wang, and Dai]{zhu2020deformable}
Xizhou Zhu, Weijie Su, Lewei Lu, Bin Li, Xiaogang Wang, and Jifeng Dai.
\newblock Deformable detr: Deformable transformers for end-to-end object detection.
\newblock \emph{arXiv preprint arXiv:2010.04159}, 2020.

\end{thebibliography}
}

\clearpage
\setcounter{page}{1}
\maketitlesupplementary

\begin{table*}[tbh]
  \centering
    \caption{Top-1 accuracy comparison for the data-rich setting. * represents the reproduced results using only the ROD datasets with the official code.}
  \label{full-data-ref}
  \resizebox{0.8\linewidth}{!}{
  \begin{tabular}{l|c|ccc|ccc|cc}
    \hline
     \multirow{2}{*}{Method} & \multirow{2}{*}{Fine-tuning} & \multicolumn{3}{c|}{RefCOCO} & \multicolumn{3}{c|}{RefCOCO+} & \multicolumn{2}{c}{RefCOCOg} \\
     &  & val & testA & testB & val & testA & testB & val & test \\
    \hline
    MAttNet \citep{yu2018mattnet}                  & w/ & 76.65 & 81.14 & 69.99 & 65.33 & 71.62 & 56.02 &  66.58 & 67.27 \\
    VGTR \citep{du2022visualgroundingtransformers}                   & w/ & 79.20 & 82.32 & 73.78 & 63.91 & 70.09 & 56.51 &  65.73 & 67.23 \\
    TransVG \citep{deng2021transvg}                & w/ & 81.02 & 82.72 & 78.35 & 64.82 & 70.70 & 56.94 & 
    68.67 & 67.73 \\
    VILLA-L \citep{gan2020large}     & w/ & 82.39 & 87.48 & 74.84 & 76.17 & 81.54 & 66.84 & 
    76.18 & 76.71 \\
    RefTR \citep{li2021referring}                    & w/ & 85.65 & 88.73 & 81.16 & 77.55 & 82.26 & 68.99 & 
    79.25 & 80.01 \\
    MDETR \citep{kamath2021mdetr}            & w/ & 86.75 & 89.58 & 81.41 & 79.52 & 84.09 & 70.62 & 
    81.64 & 80.89 \\
    DQ-DETR \citep{liu2023dq}                & w/ & 88.63 & 91.04 & 83.51 & 81.66 & 86.15 & 73.21 & 
    82.76 & 83.44 \\
    \hline
    UNINEXT* \citep{yan2023universal}         & w/  & 78.61 & 80.86 & 73.72 & 63.87 & 68.98 & 55.33 & 64.18 & 64.56 \\
    HeROD-U (ours)                          & w/  & 81.93 & 84.87 & 75.70 & 68.78 & 74.73 & 58.01 & 69.06 & 69.19 \\
    \hline
    Grounding DINO* \citep{liu2023grounding}  & w/o & 50.61 & 57.43 & 44.75 & 51.46 & 57.06 & 46.06 & 60.38 & 59.51 \\
    Grounding DINO* \citep{liu2023grounding}  & w/  & 88.86 & 91.46 & 85.73 & 80.67 & 86.99 & 72.76 & 84.66 & 84.56 \\
    HeROD-G (ours)  & w/  & \textbf{89.57} & \textbf{91.46} & \textbf{86.06} & \textbf{81.79} & \textbf{87.67} & \textbf{74.11} & \textbf{85.40} & \textbf{84.78} \\
    \hline
  \end{tabular}
  }
\end{table*}

\section{Additional Ablation Studies}
\label{sec:further-abltation}

\subsection{Data-Rich Setting.}  
\label{sec:data-rich}
Although our primary focus is label-scarce regimes, we also benchmark HeROD under full supervision to assess whether prior injection harms performance when abundant annotations are available. Results are shown in \cref{full-data-ref}. 

Compared with classical ROD models such as MAttNet \citep{yu2018mattnet}, TransVG \citep{deng2021transvg}, and RefTR \citep{li2021referring}, foundation-style detectors like MDETR \citep{kamath2021mdetr}, DQ-DETR \citep{liu2023dq}, and Grounding DINO \citep{liu2023grounding} achieve substantially higher accuracy across RefCOCO/+/g. Our HeROD variants, which simply inject reasoning priors into these backbones, match or surpass the strongest baselines. For example, HeROD-G improves over Grounding DINO by +0.7 to +1.0 points across the three datasets, while HeROD-U achieves consistent gains over UNINEXT. 

Although these margins are smaller than those in scarce-data regimes (likely reflecting the near-saturation of current detectors in full-data training) they confirm two important points: (1) reasoning priors provide measurable benefits even when abundant labels are available, and (2) our integration does not compromise performance in the data-rich case. 
This is important because explicit priors might restrict model flexibility under full supervision. Instead, HeROD maintains or slightly improves accuracy, confirming that the injected reasoning priors are complementary to learned features rather than competing with them.

\subsection{Impact of Pre-trained CLIPSeg}
\label{clipseg_impact}
To further examine the role of CLIPSeg, we compare our approach with a direct fusion baseline where CLIPSeg embeddings are simply added to Grounding DINO (see \cref{ablation-clipseg}). This naive strategy yields only a marginal gain of \textbf{+0.33} points. By contrast, our full HeROD framework achieves a much larger improvement of \textbf{+14.25} points under the same setting. 

This comparison highlights that CLIPSeg alone is insufficient for improving data-efficient ROD: its relevance maps are often coarse and do not directly influence the learning dynamics of the detector. In addition, further fusing spatial priors yields a modest gain of +0.77 over the baseline, confirming the priors are informative. The much larger gains of HeROD show that the benefit mainly comes from structured integration rather than naive fusion. In HeROD, however, CLIPSeg is reinterpreted as a visual reasoning prior and integrated systematically into the pipeline, affecting proposal ranking, prediction fusion, and the training objective. This principled integration allows CLIPSeg signals to guide both learning and inference, demonstrating that our gains come not from plugging in a pretrained model, but from explicitly embedding heuristic-inspired priors into the detection process.

\begin{table}[t]
  \centering
   \caption{Impact of CLIPSeg embedding on the RefCOCO dataset.}
  \label{ablation-clipseg}
  \resizebox{0.8\linewidth}{!}{
  \begin{tabular}{c|c}
    \hline
    Fusion of CLIPSeg Embedding & Performance \\
    \hline
    fuse w/o HeROD pipeline   & 63.99 \\
    fuse w HeROD pipeline  & \textbf{77.91} \\
    \hline
  \end{tabular}
  }
\end{table}

\subsection{Overall Impact of Spatial and Visual Priors}
To examine the contributions of spatial ($H_s$) and visual ($H_v$) reasoning priors, we conduct an ablation study over the full HeROD pipeline (\cref{addition-ablation}). Without any priors, the performance is 63.66. Adding only the spatial prior raises the score to 73.53, and integrating both spatial and visual priors further boosts performance to \textbf{77.91}. These results show that each prior contributes positively, and together they provide a complementary mechanism that substantially enhances data-efficient ROD, reinforcing the overall effectiveness of the HeROD framework.

\begin{table}[t]
  \centering
    \caption{Overall impact of different heuristic-inspired reasoning priors.}
  \label{addition-ablation}
  \begin{tabular}{c c|c}
    \hline
     $H_s$ & $H_v$ & Performance \\
    \hline
    \checkmark & \checkmark    & \textbf{77.91} \\
    \checkmark & x  & 73.53 \\
    x          & x  & 63.66 \\
    \hline
  \end{tabular}
\end{table}

\begin{table}[t]
  \centering
        \caption{Ablation studies on loss weights selection.}
  \label{loss-weight}
  \begin{tabular}{c |c}
    \hline
    $L_{conf} : L_{cls}$  & Performance \\
    \hline
    0.5  & 77.18 \\
    1.0  & \textbf{77.91} \\
    2.0  & 77.53 \\
    \hline
  \end{tabular}
\end{table}

\subsection{Loss Weight Selection}
Our final loss (\cref{eq:depr-loss}) combines three components: classification loss, bounding box loss, and heuristic confidence loss. Following standard DETR practice, we use a 1:5 ratio between classification and bounding box losses. Since the heuristic confidence term aligns more closely with classification than with box regression, we assign it the same weight as the classification loss. We validate this choice in \cref{loss-weight}, which shows that varying the relative weights has minimal impact on final performance. This suggests that HeROD’s improvements are robust to the exact weighting scheme.

\subsection{Results on Cleaned Datasets}
Recent work \cite{chen2024revisiting} identified annotation noise in RefCOCO and released a cleaned version of the dataset. While the original RefCOCO remains the most widely used ROD benchmark, we additionally evaluate HeROD on the cleaned split to test robustness. As shown in \cref{clean-refcoco}, HeROD achieves consistent improvements (e.g., +14.26 in the low-data regime), demonstrating that our approach is effective across both the original and cleaned datasets. This suggests that HeROD’s gains are not sensitive to annotation noise and generalize well under varying data conditions.

\begin{table}[t]
  \centering
        \caption{Results on the cleaned RefCOCO datasets.}
  \label{clean-refcoco}
  \begin{tabular}{l | c |c}
    \hline
     method & 0.1\% data & 2\% data \\
    \hline
    Grounding DINO & 60.57    & 68.64 \\
    HeROD           & \textbf{74.83}    & \textbf{82.42} \\
    \hline
  \end{tabular}
\end{table}

\section{Visualizations}
\label{sec:visualization}

\begin{figure}[t]
\centering
\centerline{\includegraphics[width=\columnwidth]{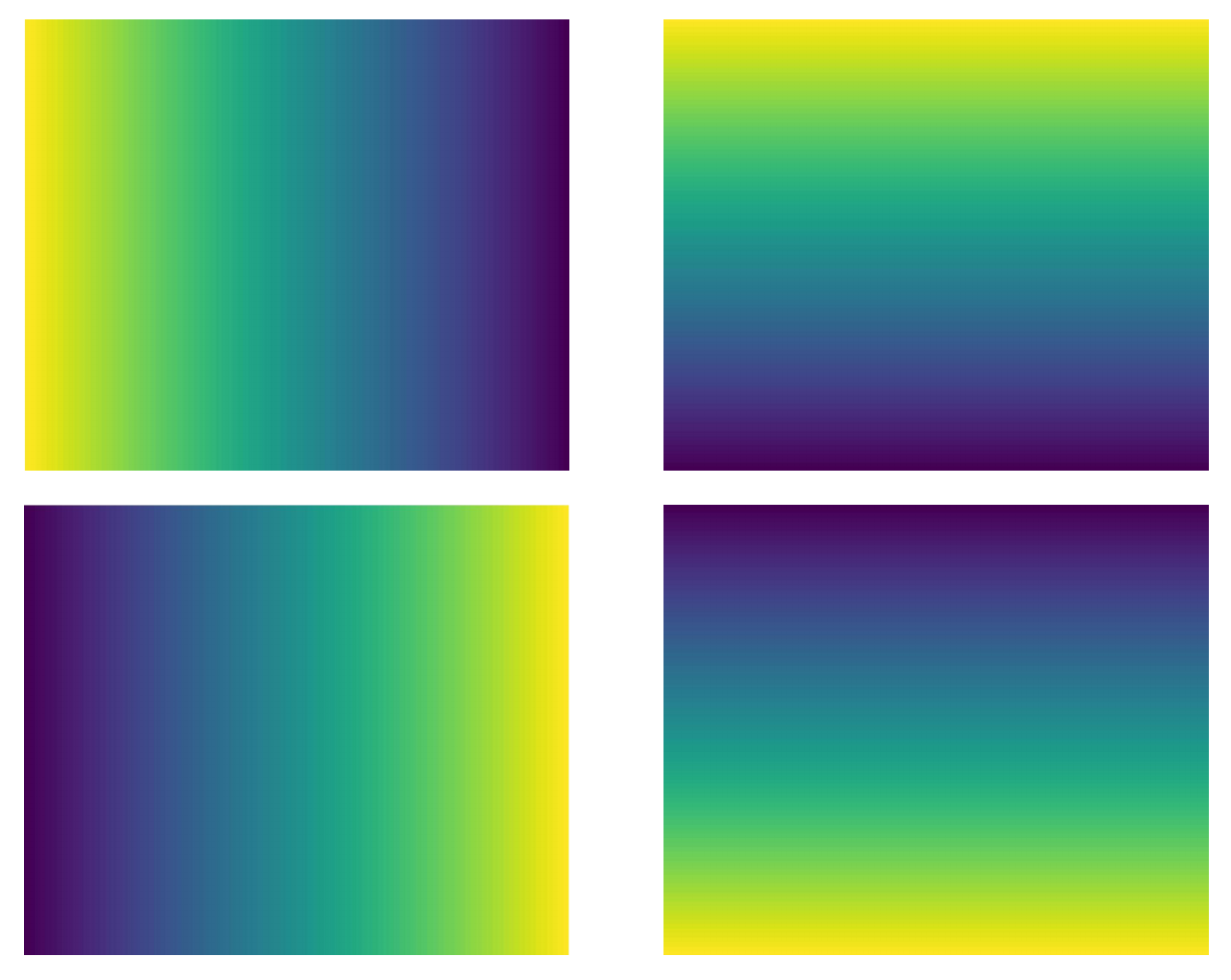}}
\caption{Visualization of four Spatial Heuristic-inspired Scoring Maps: left-related, top-related, right-related, and bottom-related (listed from the first column to the second column, from the first row to the second row). Brighter colors are used to indicate higher values, while darker colors represent lower values.}
\label{vis-scoremap}
\end{figure}

\subsection{Visualization of Spatial Prior Maps}
As described in \cref{spatial-heu}, we visualize representative spatial prior maps in \cref{vis-scoremap}. For instance, a “left” map assigns higher scores to regions near the left edge of the image, while a “bottom” map highlights areas closer to the lower boundary. For compound descriptors such as “bottom left,” we combine the corresponding maps (e.g., averaging the scores from “bottom” and “left”). These visualizations illustrate how simple positional priors provide interpretable spatial biases that can be directly injected into the detector.

\subsection{Performance across Data Volumes}
\begin{figure}[t]
\begin{center}
\centerline{\includegraphics[width=\columnwidth]{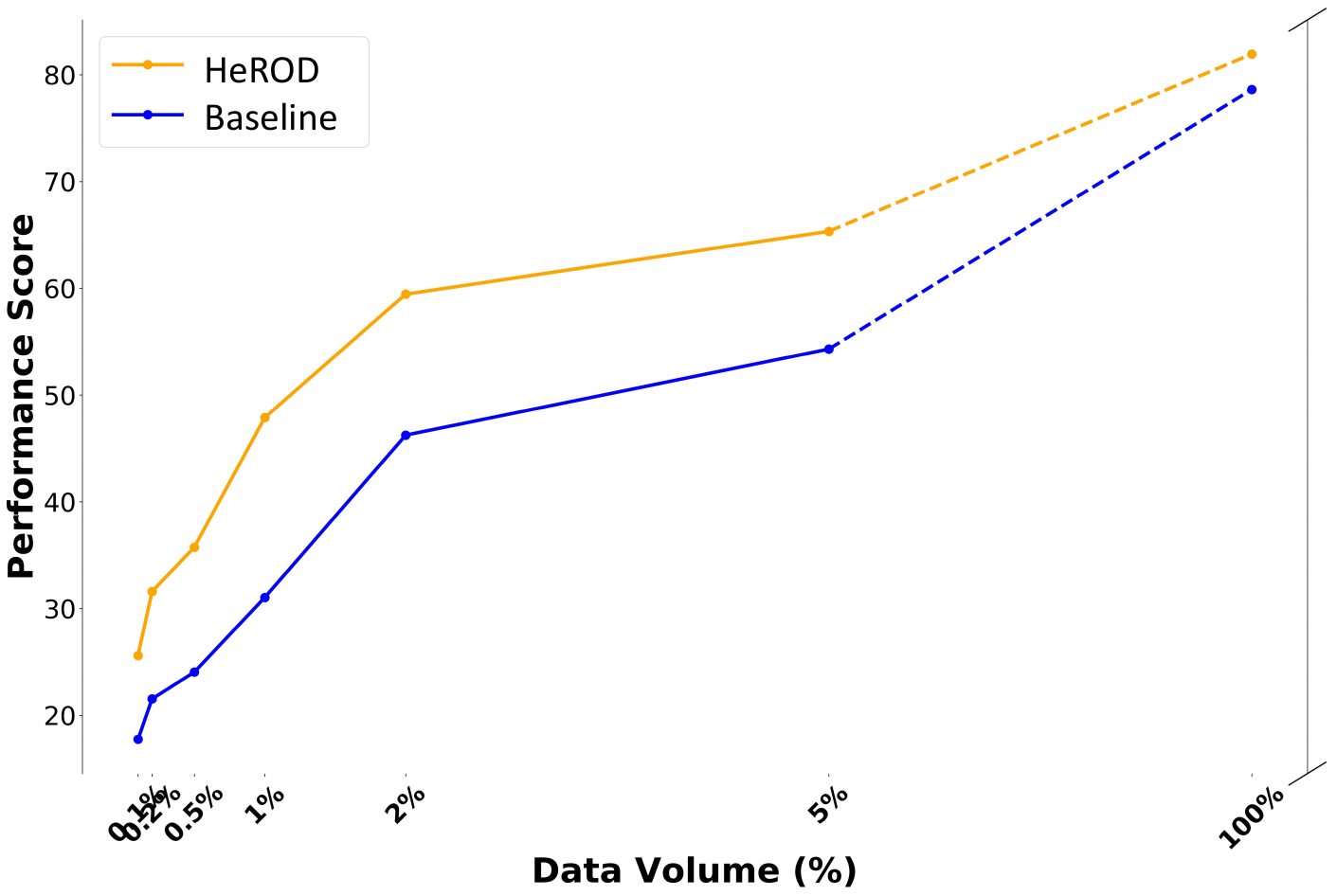}}
\caption{Visualization of performance comparison across varying data volumes between HeROD-U and the baseline UNINEXT.}
\label{curve-uninext}
\end{center}
\end{figure}

\begin{figure}[t]
\begin{center}
\centerline{\includegraphics[width=\columnwidth]{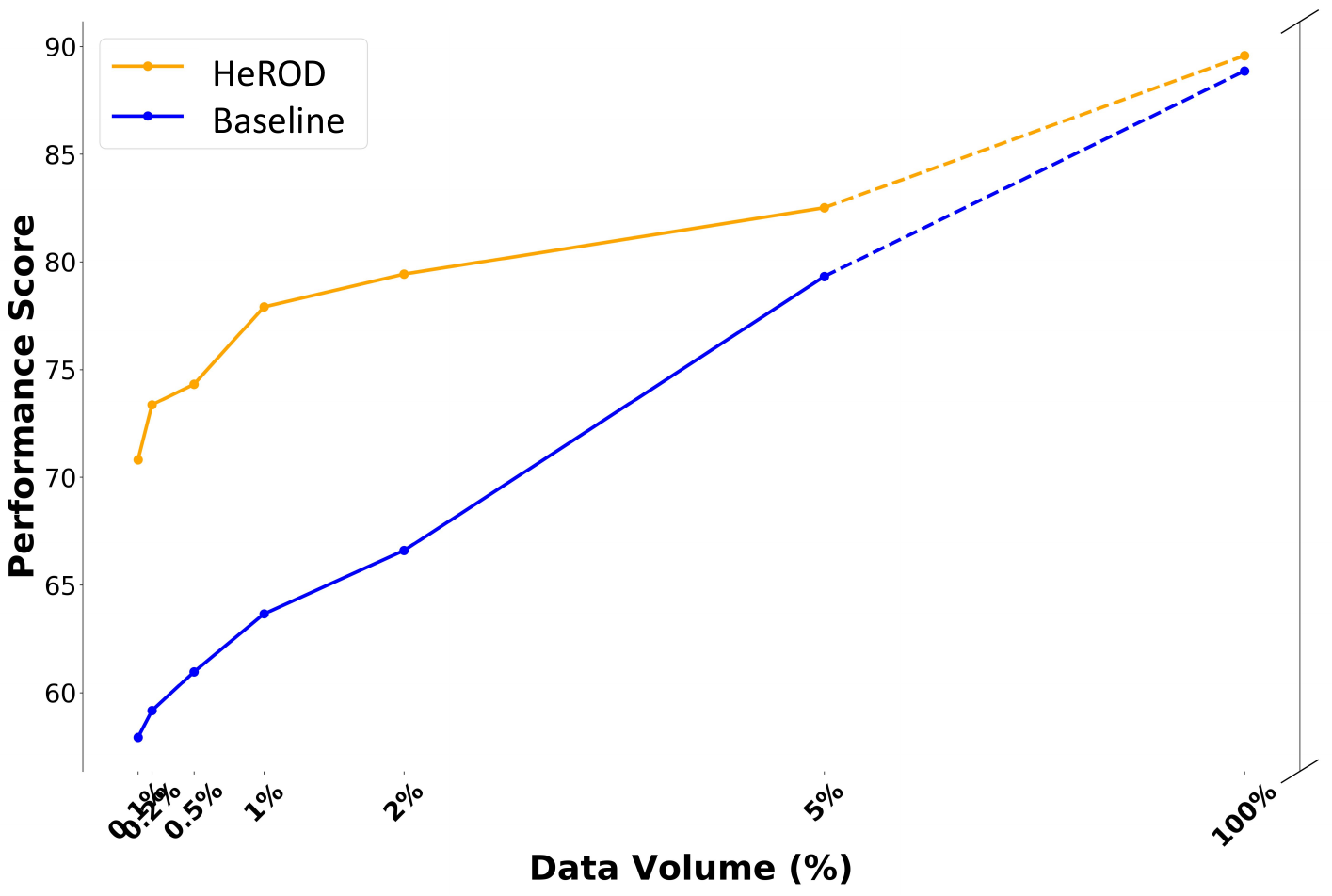}}
\caption{Visualization of performance comparison across varying data volumes between HeROD-G and the baseline Grounding DINO.}
\label{curve-gdino}
\end{center}
\end{figure}

To illustrate the effect of HeROD under varying supervision, we plot performance curves of HeROD and baseline detectors (UNINEXT \citep{yan2023universal} and Grounding DINO \citep{liu2023grounding}) across different training data volumes (\cref{curve-uninext}, \cref{curve-gdino}). The results show that HeROD yields significant improvements in low-data settings, highlighting its ability to enhance label efficiency and accelerate adaptation to the ROD task. Importantly, HeROD also generalizes well to the fully supervised regime, maintaining or improving accuracy even when sufficient annotations are available. This confirms that the injected reasoning priors are complementary to learned features across both scarce- and rich-data conditions.

\begin{figure}[t]
\begin{center}
\centerline{\includegraphics[width=\columnwidth]{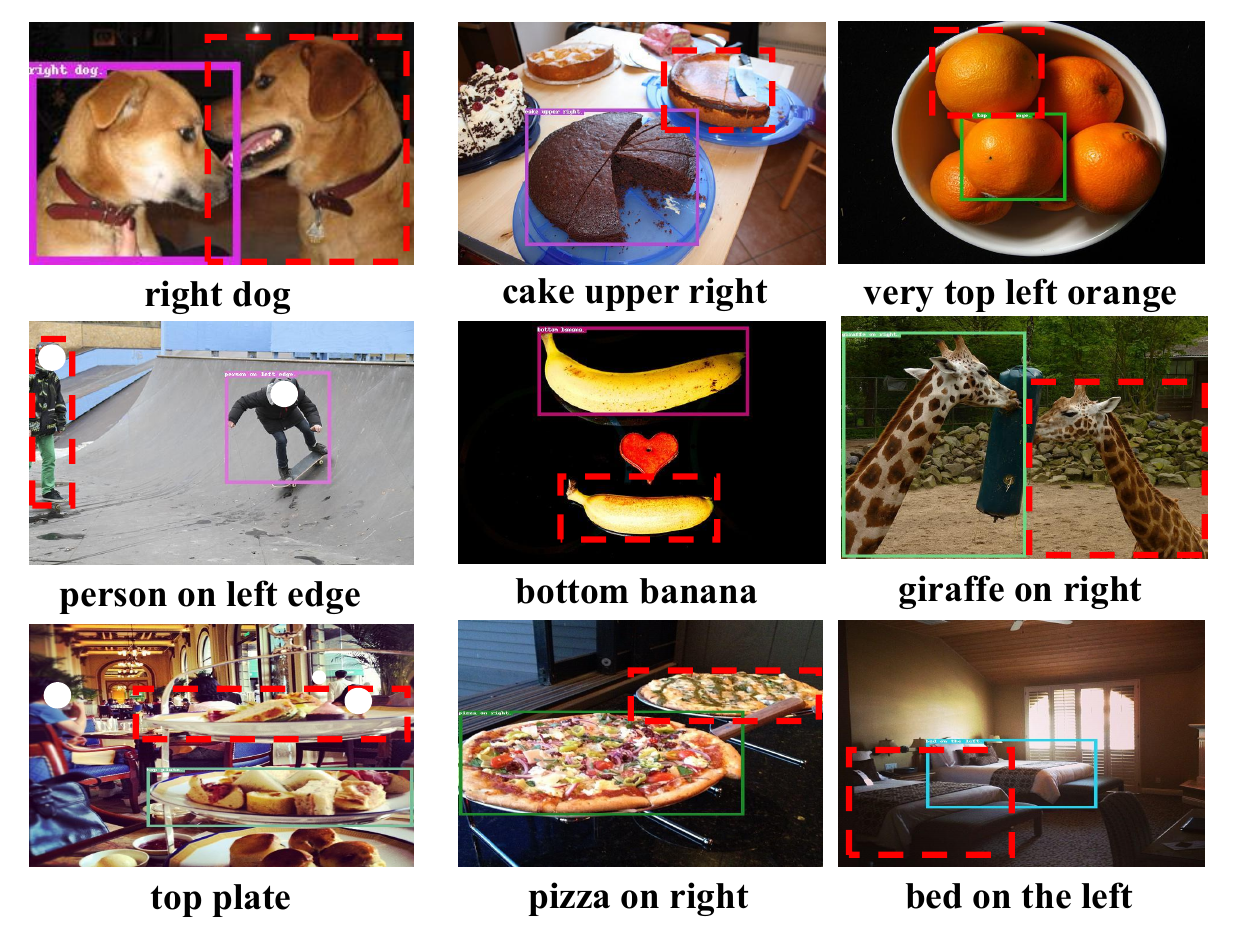}}
\caption{Predictions illustration of the pre-trained Grounding DINO. The ground truth bounding boxes are indicated by \textbf{dashed} red lines. It suggests that the vanilla grounding detection may overlook the contextual information in the text descriptions without heavy finetuning.}
\label{error-gdino}
\end{center}
\end{figure}

\subsection{Illustration of Pre-trained Detector Outputs}
As discussed in \cref{sec:intro}, we observe that the pre-trained Grounding DINO \citep{liu2023grounding} often fails to handle spatial descriptions in zero-shot evaluation. For example, when the referring phrase specifies “on the left,” the model frequently ignores the spatial cue and predicts an incorrect region. Representative outputs are shown in \cref{error-gdino}. These errors highlight the model’s insensitivity to spatial context and motivate our emphasis on injecting explicit spatial reasoning priors.

\subsection{Qualitative Examples and Discussion}
\begin{figure*}[t]
\centering
\centerline{\includegraphics[width=1.0\linewidth]{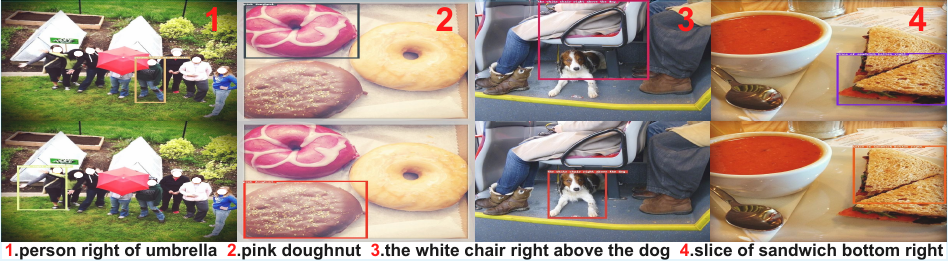}}
\caption{{\zx Qualitative examples. The top row shows correct prediction results by HeROD, the bottom row shows incorrect prediction results by the baseline. The responding referring expressions are illustrated below the figure.}}
\label{re_img}
\end{figure*}

We show qualitative examples in \cref{re_img} with different cases. In all cases, the baseline detectors fail, while our HeROD correctly localizes the target. 

For potential misleading priors, in HeROD, priors provide \emph{soft} guidance rather than hard constraints. Though absolute-direction priors can be imperfect for purely relative relations, their influence is adaptively modulated through stage-wise integration and learned fusion, allowing detector and visual evidence to down-weight misleading cues. As shown in \cref{re_img} (e.g., “person right of umbrella”), the framework remains robust under some challenging expressions. 

For complex relationships, it is true that our current spatial vocabulary is limited. This is a deliberate design choice: our goal is not to model all spatial relations, but to show that making even simple, explicit spatial priors actionable can substantially improve data efficiency under scarcity. Importantly, HeROD does not rely on spatial cues alone. Visual priors and stage-wise integration allow the framework to handle more compositional expressions in practice (e.g., “the white chair right above the dog” in \cref{re_img}). The framework is general: more expressive relational priors (e.g., relative, proximity cues) can be encoded as heatmaps and injected in the same pipeline. We view this work as establishing the foundation, with richer priors as a natural extension.

\section{Zero-Shot Setting} we also evaluate a zero-shot variant by disabling training while retaining prior injection. On RefCOCO, HeROD improves Grounding DINO from 50.61/57.43/44.75 to 59.24/64.96/53.35 (val/testA/testB), showing that it can also enhance zero-shot grounding.

\section{Latency Discussion} Baseline (without CLIPSeg) 0.362 s/iter vs. HeROD 0.435 s/iter. GPU memory rise from 4742 to 5810 MB for batch size 8. Training-time overhead is similarly small since CLIPSeg is frozen, runtime remains dominated by detector.

\section{Evaluation on Referring Image Segmentation}
Although our primary focus is on referring object detection (ROD), we also evaluate the potential of HeROD for downstream referring image segmentation (RIS). We use the high-quality bounding boxes predicted by HeROD and apply HQ-SAM \citep{ke2023segment}, a strong variant of SAM \citep{kirillov2023segment}, to generate segmentation masks without fine-tuning.

This pipeline differs from conventional zero-shot RIS approaches (\emph{e.g.}, Global-Local CLIP \citep{yu2023zero}, TAS \citep{suo2023text}, IteRPrimE \citep{wang2025iterprime}), which directly predict masks from text without intermediate box proposals. Instead, our method first localizes the referred object via text and then applies segmentation within the localized region. Despite this difference, HeROD achieves strong RIS performance (\cref{seg-ref}), showing that accurate localization serves as an effective prompt for high-quality segmentation even without mask supervision.

While not directly comparable to zero-shot RIS methods, this experiment demonstrates the broader utility of data-efficient ROD. By providing robust localization under limited data, HeROD offers a practical stepping stone toward unified vision–language segmentation frameworks that integrate heuristic-driven localization with downstream mask generation.

\begin{table*}[t]
  \centering
      \caption{Referring image segmentation results. * denotes that the paper's reported results were kept to one decimal place, and we directly replicated them here.}
  \label{seg-ref}
  \resizebox{0.8\linewidth}{!}{
  \begin{tabular}{l|ccc|ccc|cc}
    \hline
     \multirow{2}{*}{Method} & \multicolumn{3}{c|}{RefCOCO} & \multicolumn{3}{c|}{RefCOCO+} & \multicolumn{2}{c}{RefCOCOg} \\
     & val & testA & testB & val & testA & testB & val & test \\
    \hline
    Global-Local CLIP \citep{yu2023zero}   & 26.70 & 24.99 & 26.48 & 28.22 & 26.54 & 27.86 & 33.02 & 33.12 \\
    TAS \citep{suo2023text}   & 39.91 & 42.85 & 35.85 & 43.99 & 50.58 & 36.44 & 47.68 & 47.41 \\
    IteRPrimE* \citep{wang2025iterprime}    & 40.20 & 46.50 & 33.90 & 44.20 & 51.60 & 35.30 & 46.00 & 45.80 \\
    HeROD-G (ours)    & \textbf{78.21} & \textbf{80.16} & \textbf{74.30} & \textbf{71.52} & \textbf{76.86} & \textbf{63.86} & \textbf{72.52} & \textbf{72.39} \\
    \hline
  \end{tabular}
  }
\end{table*}

\section{Further Discussion of De-ROD and HeROD}
\label{Sec:dis}

\textbf{De-ROD as a benchmark.}  
Unlike convergence-acceleration methods such as Adam \citep{kingma2014adam} that improve optimization techniques, De-ROD defines a new evaluation protocol focused on data efficiency in ROD. A successful approach should achieve strong performance in severely limited-data regimes while maintaining or slightly improving results under full supervision. By emphasizing intrinsic label efficiency, De-ROD provides a realistic and stringent test bed for future ROD research, especially in scenarios where large-scale annotations are infeasible. This benchmark highlights both the urgency and the value of designing ROD methods that require fewer labels yet deliver competitive accuracy.

\textbf{Scope of spatial and visual priors.}  
Our current spatial priors ($H_s$) are limited to cardinal and compositional directions, and thus cannot capture more complex contextual relations. This limitation is partially mitigated by the visual prior ($H_v$), which leverages CLIPSeg to align phrase semantics with image regions. Importantly, our approach goes well beyond simply adding CLIPSeg scores or ensembling with a grounding detector. CLIPSeg maps alone are coarse and do not influence detector learning; as shown in \cref{clipseg_impact}, naïve fusion yields only marginal gains. In HeROD, CLIPSeg signals are reinterpreted as visual reasoning priors and integrated systematically into proposal ranking, prediction fusion, and the training objective, which enables them to directly shape learning. While CLIPSeg is used here for convenience, other pretrained models that provide local alignment signals could be employed in the same role.

{\zx \textbf{Scope and limitation.} Our study targets data-scarce referring object detection on standard benchmarks, under the common assumption in modern vision–language research that pretrained foundation models are available. As such, HeROD is most applicable to real-world settings like robotics, where language-guided localization is needed but dense labeling is costly. A limitation is that our semantic prior relies on the availability of suitable pretrained models; in niche domains (e.g., medical imaging), effective deployment may require domain-specific foundation models/priors and datasets to produce reliable heatmaps. The framework itself is general, and can be adapted when appropriate domain foundations are established.}

\textbf{Novelty and extensibility.}  
The novelty of HeROD lies in the systematic integration of spatial and visual reasoning priors into DETR-based detectors (\cref{HeROD-pipeline}). Unlike prior work that relies solely on implicit feature learning, HeROD explicitly injects interpretable priors into multiple stages of the pipeline, reducing dependence on large labeled datasets and improving cross-modal alignment. Although this work focuses on spatial and visual priors, the framework is extensible: additional priors, such as structural patterns, domain expertise, or medical knowledge, could be incorporated to further enhance data-efficient referring object detection. {\zx For instance, depth-based terms may be feasible within HeROD. Depth cues from a depth estimator can be converted into text-conditioned depth priors and injected as reasoning signals in the same pipeline.}

\end{document}